# Machine Learning for Exploring Spatial Affordance Patterns

by

**Boyana Buyuklieva**

September 2015

A Dissertation submitted in part fulfilment of the
Degree of Master in Adaptive Architecture and Computation
Advisor: Dr Martin Zaltz Austwick
Technical advisor: Seda Zirek

The Bartlett School of Graduate Studies
University College London

I, Boyana Buyuklieva, confirm that the work presented in this thesis is my own. Where information has been derived from other sources, I confirm that this has been indicated in the thesis.

# Abstract

This dissertation uses supervised and unsupervised data mining techniques to analyse office floor plans in an attempt to gain a better understanding of their geometry-to-function relationship. This question was deemed relevant after a background review of the state-of-the-art in automated floorplan generation tools showed that such tools have been prototyped since the 1960s, but their search space is ill-informed because there are few formalisms to describe spatial affordance. To show and evaluate the relationship of geometry and use, data from visual graph analysis were used to train three supervised learners and compare these to a baseline accuracy established with a ZeroR classifier. This showed that for the office dataset examined, visual mean depth and integration are most tightly linked to usage and that the supervised learning algorithm J48 can correctly predict class performance on unseen examples to up to 79.5%. The thesis also includes an evaluation of the layout case studies with unsupervised learners, which showed that use could not be immediately reverse-engineered based solemnly on the VGA information to achieve a strong cluster-to-class evaluation.

# Content

# Acknowledgements

I would like to thank Dr. Martin Austwick for this advice at every step of this research and for the encouragement he provided for the pursuit of my interests in this dissertation. Without this, I doubt I could have had the confidence to venture out and learn so much. I am also very grateful to Dr. Sean Hanna for the inspiring ideas and projects he introduced every Monday morning that sparked my interest in data mining and machine learning in the first place. Finally I would like to express my gratitude to SpaceLab for supporting this dissertation by providing the office case studies.

# Table of Figures

# Introduction

## 1.1 Motivation

The interest of this thesis is the floor plan at an architectural scale. The initial intent of the work was to create a layout generating tool based on an expert-system trained on real examples that could suggest function/usage spaces very much like an auto-predict. As the auto-correct reminds us about spelling thereby helping us bring content faster without imposing its own intent, so could perhaps a tool be developed to suggest space allocation to the architect without removing him from the design process. However, as research for this project began it was confirmed that architecture as a field knows too little about the grammar of its own language as there are few design formalisms (Merrell et al, 2011, p.3) that can be implemented to make such a system. Therefore the focus of this thesis shifted into the study of spatial affordance.

## 1.2 Background

### 1.2.1 Review of Automated Layout Tools

Architectural layout generation is the act of allocating usage spaces with topological and geometric relationships according to some design criteria (based on Jo and Gero, 1998) and within some site boundary. This process can be divided into three phases: the planning, analysis and synthesis stage (Rodrigues 2014, p.7). The first is the arrangement of spaces according to some abstract design intent or concept. This is followed by a gathering of diverse functional requirements and constraints. Finally, the synthesis stage is an iterative trial and error process where different layout configurations are draw out to accommodate the goals of the first two phases. The design of an architectural layout is a very time-consuming and repetitive task, prone to human error. Despite this, it is still done manually because it requires flexibility and creative problem-solving (Rodrigues, 2014, p.17). Nonetheless, there have been prototypes developed for automating the layout generation process. This section will examine some recent such projects.

A very recent example (2014) of automated floor plan design is given by E. Rodrigues. His doctoral thesis proposes a tool for the early design phase, specifically the space planning phase, which generates different energy-efficient floor plan configurations. In his approach, the architect's role is to input geometric requirement into a complex matrix, express a compactness preference, allow the algorithm to produce a number of resulting configurations and, finally, choose from these and refine them. Although the system succeeds in replicating results made by hand and positions itself earlier in the design phase (Rodrigues, 2014, p.143), it puts the architect's own ideas in the background as the search space of the designer has little direct influence over the search space of the algorithm. Rodrigues' model envisions the role of the architect not as a creator, but rather as a curator – an approach that has been discussed thoroughly in academia as early as the 1960s but with little practical acceptance and implementation (Liggett 2000). A different example of the generation of automatic, yet convincing buildings is Duarte's use of discursive grammar to produce houses in the same style as Siza's Malagueira Houses. These were made using a series of questions to a user and a set of shape grammars derived from real projects from Alvaro Siza's archives (Duarte, 2001). This approach is different from the other projects for automatic layout generation outlined because it applies user input to a search space that is limited to configurations of space types according to a set of learnt rules. The value of this work is that it goes beyond the *optimization problem* by encoding and applying empirical patterns learnt from the architect's archives to create user-adapted buildings; ones that Siza himself could not distinguish from his own (Duarte, 2001, p. 17).

Another example is provided by Merrell et al (2011). They present a method for applying machine learning techniques for the automatic generation of residential layouts and houses for computer graphics applications. Three Bayesian neural network were trained to create one, two and three story houses. The total number of training instances used was 120 manually-encoded residential cases, in which every node is a room from the architectural program. Their work shows that expert systems in architecture are possible because their system creates 'visually plausible building layouts' (Merrell et al., 2011, p.9) only from parameters taken from a catalogue of floor plans. Although Merrell et al. manage to generate whole residential buildings, they point out that their method is "idealized and does not take into account the myriad of site-specific and client-specific factors that are considered by architects" (Merrell et al., 2011, p.9). This project shows that floorplans have inherent information in them which, when studied, allows for non-architects to re-create residential houses. However, it also points out that despite the potency of automation, the role of architects is still important because of their natural ability to adapt to physical and social context more easily that the current state-of-the-art in layout generation.

Homayouni (2006) describes three categories of computational design synthesis methods with a specific focus on automated space planning (see table 1). Procedural methods are those that iterate all possible solutions but they have the obvious setback that as the layout becomes more complex, the number of configurations grows exponentially and it becomes impossible for the architect to invest the time and energy to review such solutions. These methods are often coupled with stochastic procedures to generate unexpected outcomes and evaluate a larger search space with less iterations. Homayouni refers to these as 'evolutionary methods'. This coupling of procedural searching with evolutionary methods has been explored in multiple prototypes, including Rodrigues'. The heuristic or analogical methods, to which Merrell et al and Durate belong have received less attention and are therefore underexplored in comparison. It can be argued that these are more difficult to implement because of the amount of previous case studies required for their basis.

| **Approach to Automated Space Planning** | **Description** | **Sub-Categories** |
|---|---|---|
| Procedural | Enumerating all possible solutions in the search space | Space Allocation: enumeration with some rationale/ ranking (ex. Desired distance ) |
| | | Constraint Satisfaction: space allocation with additional criteria (ex. solar exposure) |
| Heuristic / Analogical | Empirically informed search space based on analogies to other built projects. | Case Based Methods: recycling of previous layouts by re-modifying these to satisfy the new situation |
| | | Expert Systems: applying rules learnt from previous layouts databases to generate new ones |
| | | Shape grammars: expert systems in which the rules are geometric modules that are re-arranged |
| Evolutionary | Heuristic or procedural methods combined with a stochastic exploration of the search space | Genetic algorithms/ programming: 'breeding' of successively better solutions based on some predefined fitness criterion |

*Table 1 - Approaches to Automated Space Planning based on Homayouni with brief overview of each*

The examples mentioned in this section lead to the conclusion that automated layout generating tools are available, but they are of insufficient quality to be employed as the norm. Architects have demonstrated open-mindedness towards software tools to improve their work flow (Homayouni,2006, p.24; Kanters et al. 2014, p.725), but semi and completely automated layout tools have not fared well. They do not provide a good user experience because the current state-of-the-art tools reduce the architect to a curator, and also because most present prototypes require a long runtime to generate results that will need to be further re-worked manually. Briefly stated, failure of these tools lies in their interaction quality and unreasonable search space. This thesis will focus on the latter problem using techniques borrowed from computer science.

### 1.2.2 Data Mining

Data mining is an experimental science that aims to extract implicit information and relationships from a dataset by combining statistics and computation. At its core are a set of machine learning algorithms, which can be generally summarised as falling into one of three main categories: supervised learners, where there is a correct output to strive towards, unsupervised ones, where the learner has to figure out a structure based on unlabelled data and reinforcement learners, which do not have labelled data but create an output based on user-provided rewards/punishment (Duda et al., 2001). Given a database, a learner can output new information under the form of either a regression or a classification. Regression is useful for estimating or predicting the numeric value of a new instance; classification for sorting the instance to some class membership. In either case, the algorithms applied have some assumption or knowledge beyond the dataset given to form their prediction, therefore every technique necessarily has some bias (Nilson, 1998, p.80). It has been shown through the "No Free Lunch Theorem" that there can exist no universally best algorithm that is optimal for any dataset (Ho and Pepyne, 2001) – nonetheless the performance of any algorithm is directly dependent on its training experience (Mitchell, 1997, p.17). Regardless of the learner, the provided inputs are referred to as examples, or *instances* of the dataset. These have certain characteristics associated to them, referred to as *attributes*, and can be mainly of two types: nominal (discrete/categorical) and/or numeric (continuous). The instances and their attributes make up the training experience for any learning scheme.

### 1.2.3 Overview of Informed Design

In manual layout design, architects rely mainly on rule-of-thumb approaches and their past experiences to explore possible solution spaces (Kanters et al, 2014; Rondrigues 2014; Merrel et al 2011; Duartes 2001). Although some formalisms and guidelines have been proposed, "few are nearly universal" (Merrell et al, 2011, p.3) and only recently has there been an increased interest in studies to promote data-driven, informed design. Examples of work in this direction are the studies of Koutsolampros et al (2015) and Sailer et al (2012) in the field of office layout design. These studies are not limited to a single style, but rather use behavioral and space syntax measurements as a foundation with which to quantify and compare different floor plan configurations. Similarly, R. Pachilova (2015) has looked at patterns and clusters of spatial arrangements, specifically for hospital building types. These studies are interested in socio-spatial patterns, but also in geometric measures such as visibility, connectivity and other space syntax measurements, which will be discussed in the following sections. The research conducted by Sailer, Pachilova and Koutsalompros et al is a step closer towards a more heuristic, informed design approach because it attempts to derive implicit, numerical patterns from multiple real world examples to be applied to future projects. This is especially valuable for the development of automated, or semi-automated floor layout generation because it can empirically inform the search space of such algorithms, thereby producing more plausible results in a shorter amount of time. However, as the authors point out in their papers, the studies have been limited due to incomplete, in the case of Pachilova, and non-systematic, in the case of the other two, databases.

Another approach that deals specifically with the classification of floor plans using data mining, more precisely self-organising maps (SOM), can be found in the work of Derix and Jaganath (2014). Here the SOMs are tested as a mechanism to classify spatial attributes of a layout to find new emergent spatial patterns. Derix and Jaganath are interested in implicit morphologies of spatial configurations and their perception. Their research seeks to analyse the affordances of built spaces and also to inform the early design phase. Although their work was mainly carried out as part of a study in spatial resilience and infrastructure security with only three case studies, their methodology is very relevant.

They discretise the layouts and attach conventional or extended space syntax measurement values to them. This is interesting because it makes layouts comparable in a symbolic way. Also, it provides a framework for modular collaboration because different interest groups or engineers can attach different numerical values for the partitions. The value of this work lies in the idea that typologies can be derived rather than assigned based on associations between geometric properties (mainly space syntax measurements) without referring to the configuration's pre-assigned functional type. Therefore a typology is presented as a weighted combination of attributes which account for a particular "use affordance type" (Derix and Jaganath, 2014, p. 3).

### 1.2.4 Overview of Space Syntax

Space syntax is a collection of theories and complementary space analysis techniques that were presented by Hillier and Hanson in the 1980s (Hillier & Hanson, 1984). At its core is the idea that space and society, or in other terms, geometry and use, are tightly interwoven. They argue that buildings are not usual artefacts because they create and organize spaces. The configuration of these spaces is important because they give the layout meaning. Meaning here can be understood as Dorish's idea of embodied meaning: "a set of relationships between affordances and the environment" (Tuner, 2008). Therefore, space in this context refers to creating different opportunities for social interaction and use (Hillier and Hanson, 1984; Hillier, 2007). The analytical techniques in space syntax measure spatial relationships by graph theory: plans of spaces can be represented as a discrete graph with nodes for which different relative values can be computed (Al-Sayed, Alisdair and Hillier, 2013, p.27). These values are interesting because they have been shown to correlate with certain use and perception patterns. For example, there are studies that indicate that integration (explained later in section 2.1.2) has a strong correlation with pedestrian movement and is indicative of social encounter rates (Turner, 2001, p.14).

All space syntax measurements can be described as a global or a local value. Global values are measurements that are calculated for a node with regard to all other nodes in the system, whilst local values examine a node with regard only to the nodes directly connected to it (Turner, 2004, p.14).The standard tool for this particular type of spatial analysis is depthmapX (Spacesyntax.net, 2015) which deals exclusively with graph analysis (Turner, 2004, p.18). This thesis will focus on the visibility measurements taken from space syntax. These are important for spatial cognition and the resulting use of space. The visual perception of a layout is the most intuitive and has been shown to predict how accessible spaces afford movement (Al-Sayed, Alisdair and Hillier, 2013, p.27). The correlation between movement and visibility is best explained through Gibson's idea of natural vision: "When no constraints are put on the visual system, we look around, walk up to something interesting and move around it so as to see it from all sides, and go from one vista to another. That is natural vision..." (Gibson, 1986, p. xiii). Visibility measurements can be described as either physical, taking into account the metric distance, or visual, describing cognitive distance. For the latter, the view shed or isovist (Benedikt, 1979) is important. This refers to the space directly accessible to our visual system from a position in the layout. In space syntax, the isovist can be represented as visibility for which each node can be connected to another inter-visible node (Al-Sayed, Alisdair and Hillier, 2013, p.27). The exact measurements used in this thesis can be found in section 2.1.2.

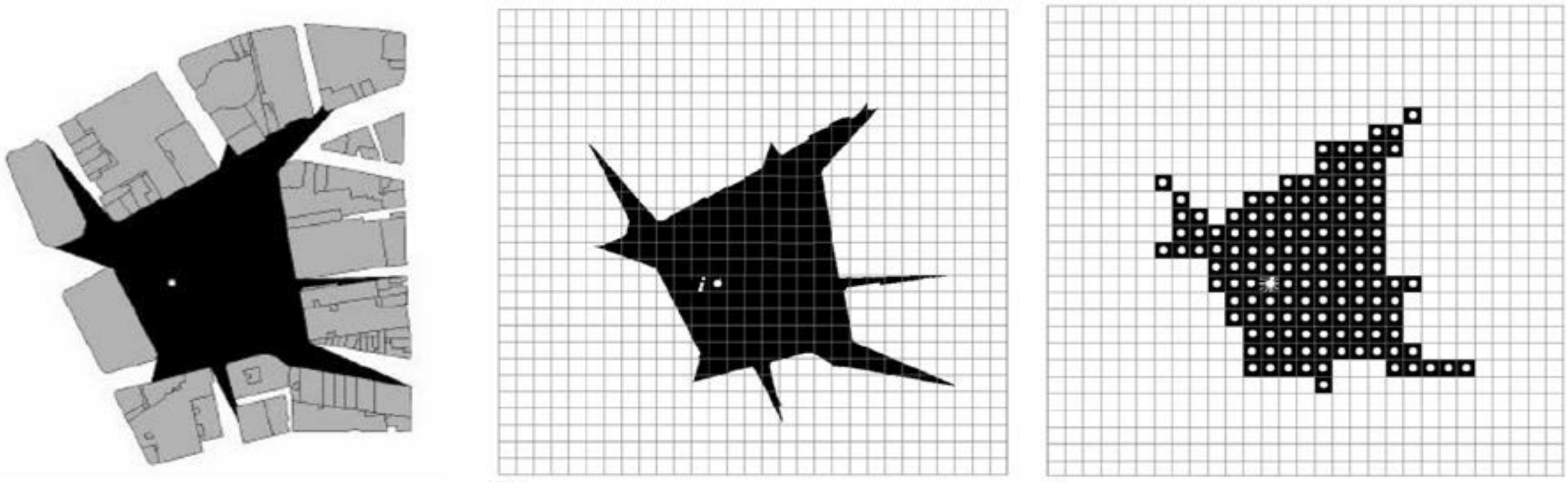

(a) (b) (c)

*Figure 1 (Based on Batty, 2001, p.126): (a) Isovist polygon generated for point i (b)/(c) Isovist polygon as nodes in a grid*

### 1.3 Goals and Outline

This dissertation is interested in borrowing data mining techniques to examine use affordances in office floor plans. It presents two hypothesis. The primary claim is that there is a link between geometry and function, therefore a learning algorithm can be trained to recognise what a space is used for based on its space syntax analysis. The secondary hypothesis holds that use can be reverse-engineered with an unsupervised learning algorithms to achieve a strong cluster-to-class evaluation.

Section 2.1 of the methodology will describe the office dataset acquired from SpaceLab in regard to its VGA. Thereafter, an introduction to data mining and how the layouts are represented in this context is provided. An overview of the experiments that were undertaken to show how the different geometric measurements relate to the usage can be found in 2.2.1. The results section shows the outcomes of all relevant experiments. Following the first test, and taking into account the unusual span of the data, section 3.2 provides a closer study of the specific space syntax measurements for calculating normalised mean depth. This is valuable because to the best of the author's knowledge, the literature on when to implement which integration value is ambiguous and no study of these measurement on such a scale has been done, especially not on layout interiors.

# Methodology

## 2.1 The Office Dataset

### 2.1.1 Source and Overview

To explore the idea of spatial affordance based on geometric data and learn rules from previous layouts, this thesis explored 14 case studies of offices in the UK as provided by SpaceLab, a London-based architectural and consultancy service. The case studies provided a total of 36 layouts. Each layout consists of a CAD drawing (.dxf) of the accommodation and its individual visibility analysis conducted in depthmapX. Files in the .dxf format can be read as .txt files comprised of five sections: header, classes, blocks, tables and entities. Each of these sections contains tagged default or user-specific data associated with a specific group code (Autodesk.com, 2015). Although open .dxf to .csv converters can be found (Ransen, 2015), for the purpose of the thesis, the .dxf data was converted with the consultancy's in-house tool. This was advantageous because it allowed for direct coupling of the .dxf generated text file with the automatically generated .csv file from depthmapX. Following this procedure a text file (.cvs) for every floor plan was extracted. This file represents the floor plan as a discrete grid made up of .45m x .45m units (the resolution of the grid depended directly on the resolution of the depthmapX analysis). Each grid contained the following 17 values, which were either layout specific or measurements from the VGA analysis (see section 3.2):

Layout specific measurements

- Ref : Unique ID of the unit
- X : Relative x-position of the unit
- Y : Relative y-position of the unit
- Accommodation / Accommodation Poly ID Use function assigned to the unit \ Polygon ID of this
- Team / Team Poly ID: Team that occupies the polygon of the unit \ Polygon ID of this

### 2.1.2 Definitions of VGA Measures

Connectivity

A local, visual measure, which shows how many other nodes can be seen directly from a certain node. (Klarqvist, 1993, p.2)

Visual Node Count:

Number of nodes in the layout, a physical measurement.

Visual Mean Depth:

A global, visual measure that can be calculated for every node. It looks at the average of the shortest paths (the ones with the smallest number of turns) between a node and every other node in the system and is measured in steps. (Turner, 2004, p.14) Mean depth is indicative of topological distance.

Point First Moment:

A local, physical measure that relates to connectivity. Point first moment is defined by the sum (not average) of the steps from one node to every other visible node in the system (Al-Sayed, Alisdair and Hillier, 2013, p.33). This measure is indicative of how far it can be seen in a space.

Point Second Moment:

This is the standard deviation of the connectivity value. This local, physical measure is relevant for estimating the perimeter of a space. For example, compact spaces would be expected to have a low point second moment and high values are more likely to occur in an elongated corridor.

Visual Entropy (Point depth Entropy)

It is a global measure that indicates how orderly a place is by giving the most compact description of how depth values, d, can change around a node. It examines the frequency distribution of the depths, or otherwise stated: the ease of moving from a space with one depth value to another with a different depth value. As such it can be considered a visual measure. It is useful in predicting how ordered a system is as perceived from a certain node since it reflects the average number of turns required for any journey within the system (Tahar and Brown, 2003, p.6). Low entropy measures correspond to low disorder and an asymmetric depth distribution, which indicates that it is easy to navigate in this space. High values correspond to high disorder and a visual depth that is evenly distributed, which means that it is harder to move about (Turner, 2004:21, Turner 2001, p.8).

Point depth entropy is based on Shannon's idea of entropy as used in information theory (Turner, 2004, p. 21). For every node in the layout, the frequency of other nodes with depth *d* to it is taken into account. This frequency is given by $p_d$.

$$Point\ Depth\ Entropy = \sum_{d=1}^{d_{max}} -p_d \log p_d$$

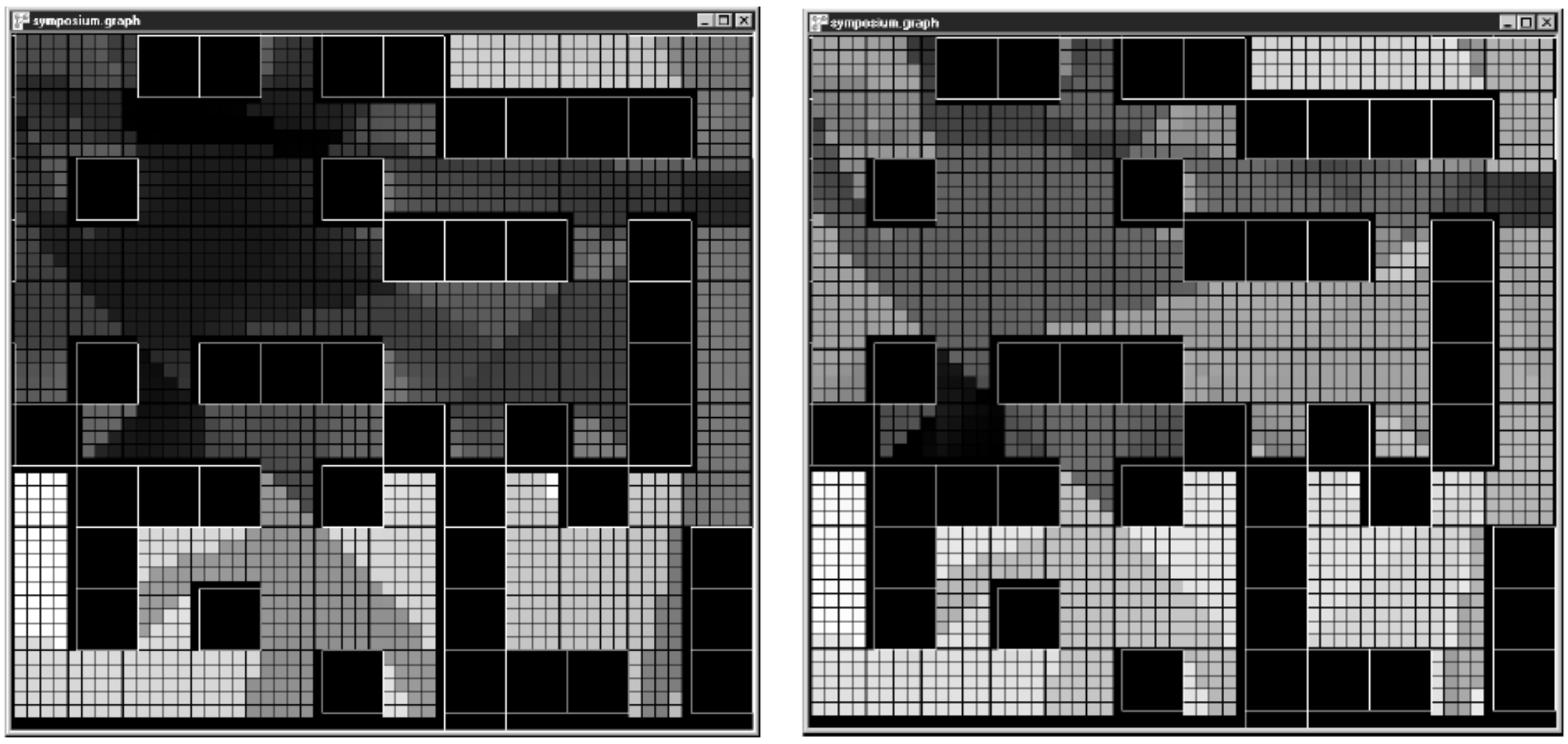

*Figure 2 (Turner 2001, p.8): The left image depicts the layout colored by mean depth; the right image depicts the same layout with coloring for point depth entropy. In each image case, the higher values are brighter colors. Point depth entropy is useful in eliminating the area dependency bias of the VGA towards open spaces in order to concentrate on the visual accessibility of a node in relation to all other nodes (Turner 2001, p.8)*

Visual Relativized Entropy

An extension to point depth entropy that gives the probability of splitting at the next node: "the frequency with regard to the expected frequency of locations at each depth" (Turner 2001, p.8). It assumes a Poisson's distribution, by including the Poisson's probability mass function ($q_d$).Thereby this measure 'relativizes' the entropy calculation by taking into account the expected, average mean depth ($L_i$).

$$Relative\ Point\ Depth\ Entropy = \sum_{d=1}^{d_{max}} -p_d \log \frac{p_d}{q_d}$$

$$\text{for} \quad q_d = \frac{L_i^d}{d!} e^{-L_i}$$

Integration

A global, physical measure that is mean depth normalized against some established graph with an equivalent amount of nodes. The established graph can be: a diamond root graph [HH], a corner-of-a-grid graph [P-Value] or a bipartite graph [Tekl] (these measures are described in more detail in section 5.2). The integration values can describe spaces as segregated or integrated. The latter are spaces claimed to afford a high probability of social encounter (Al-Sayed, Alisdair and Hillier, 2013, p.13).

### 2.1.3 Grouping Usage Classes

The provided data divided spaces into 12 categories, based on their general use. Missing values were examined for patterns and it was found that these can best be described by their own group of spaces that serve for secondary circulation, as these were movement corridors that lead into the office spaces without being a part of them directly (see appendix A for more details). Following a grouping by similarity, the 39 labels from SpaceLab were initially generalised into the 11 specific categories (table 2). These class values were used for all supervised studies, despite the strongly disproportional distribution of classes such as G6, G5 and G8.

| Class | Total instances in class | Percent of all |
|---|---|---|
| **G1: Open Workspaces**<br>Open plan desk area, alternative open work spots near main workspace and other alternative workspaces | 65140 | 36.9% |
| **G2: Cellular Workspaces**<br>Tables, sofas or booths in an enclosed space, usually offices of executives that receive guests and therefore need more privacy or offices of small teams. These can be alternative spots close to the main work area or further away with no visual connection. Included are also (semi) enclosed points with books/magazines and quiet rooms. | 17371 | 9.8% |
| **G3: Meeting Spaces**<br>These include bookable and non-bookable enclosed spaces, flexible alternative spaces and videoconferencing rooms. What unifies these rooms is the fact that they are shared enclosed and allow for temporary use. | 21443 | 12.2% |
| **G4: Storage Facilities**<br>All storage types are considered – any height or enclosure. Included are also server and plant rooms. | 8298 | 4.7% |
| **G5: Kitchen/Tea Seating**<br>These include canteens and their seating areas. Also open and closed tea spaces. | 8137 | 4.6% |
| **G6: Kitchen/Tea Service**<br>These include catering kitchens, canteen serving areas and externally run tea/coffee/vending stations. | 2264 | 1.3% |
| **G7: Print/Copy Facilities**<br>Any space used for printing, copying and in some cases, cutting. These includes a single machine or dedicated room. | 1829 | 1.0% |
| **G8: Reception**<br>This includes the reception, the small circulation space around it and any related furniture. Pigeonholes and other functional spaces that cannot be used for work are also considered here. | 3193 | 1.8% |
| **G9: Extra Facilities**<br>In the case studies examined there are a gym and a nursery. But they can be generalised to any miscellaneous, permanent facility, not directly linked to the activities in the office. | 370 | 0.2% |
| **G10: Primary Circulation**<br>Movement corridors within or directly adjacent and leading to the open plan office space. | 35282 | 20.0% |
| **G11: Secondary Circulation**<br>Movement corridors or spaces that do not directly lead to the open plan work spaces. | 9544 | 5.4% |
| **EXCLUDE:** These include spaces that don't relate to the offices or storage below eye sight (furniture). | 3524 | 2.0% |

*Table 2: Description of usage classes in the case studies (A more detailed description of usage types and their original labelling can be found in appendix A.)*

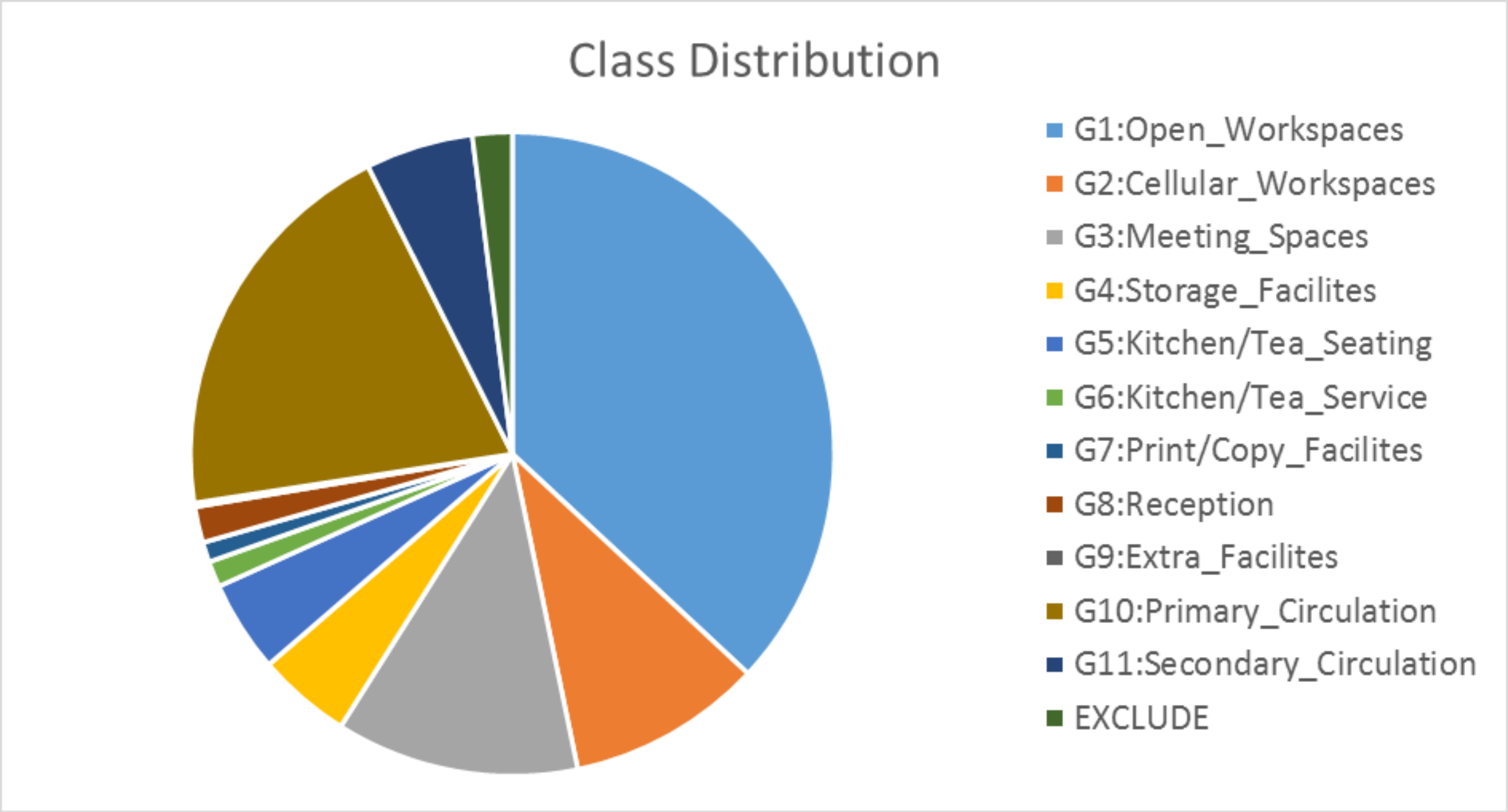

*Figure 3: Overview of the initial classes and their proportions in the dataset*

## 2.2 Experimental Setup

This thesis uses supervised and unsupervised learning algorithms to evaluate the floor plans for usage patterns by means of classification. In doing so it also tests the value of the space syntax measurements, specifically those relating to integration. For these purposes, an instance taken from the dataset of 36 layouts has been defined as a node in the VGA graph. Although Derix and Jagannath use room-like partitions based on largest convex space, they argue that this is a limitation in their work because "inherent trends in spatial values across the open plan spaces of the layouts [become] camouflaged by the pre-determined divisions of the convex spaces" (Derix & Jagannath 2014, p.212). Considering the instances as larger, room-sized polygons would bias the data because, although a simple rectangular room could be adequately described in this way, this method would necessarily require some strong assumptions to deal with irregular rooms. These assumptions are problematic because multiple functions can accommodate an open space, especially in office buildings.

The experiments in this research have been done in WEKA (3.6/3.7). WEKA 3.7 is the current developer version of the java-based, open-source data mining workbench from the Department of Computer Science at the University of Waikato. This choice was made for several reasons, the most obvious of which is that data mining requires a combination of breadth and depth. Depending on the structure of the data different learners will necessarily perform better than others, regardless of data preparation, therefore it is useful to test a dataset on many learners. WEKA's library is good for these purposes, because it has a good breadth but is also flexible enough to be tweaked to suit the exact need of the data at hand. Another reason for the implementation of this particular library is the fact that its source code is openly available for private use and can easily be extended for commercial purposes (Weka.wikispaces.com, 2015). This means that results obtained from it can be integrated into a larger application of any relevant tool, including one for heuristic automated or semi-automated floor plan generation. Finally, testing the dataset on this tool makes the results of this research capable of being replicated and its methodology can be directly re-implemented for any other datasets.

### 2.2.1 Quality of the Training Experience

#### *Training Schemes*

A learner's ability to produce new information can be explained as "searching through a space of possible hypotheses to find the hypothesis that best fits the available training examples" (Mitchell,1997, p.18). A good learner is one that is able to perform well on new data, beyond its testing set or, otherwise stated, one which generalises well instead of fitting tightly to its training set. Overfitting can be an issue for any learning scheme, but, it is especially problematic where there are too many hypotheses imposed on a small training set. Care has been taken to keep the number of instances large (176 365 instances) to provide an adequate training set.

There are several methods to test how well a model generalises, the most basic of which is the *holdout* method, most commonly used as *stratified holdout*, where random sampling is involved. This method simply divides the input attributes into a training set and test set, typically 2/3 (Nilson,1889, p.83; Witten, Frank & Hall, 2001, p.152). There is a trade-off to this method, especially when working with limited data because reducing the size of the training set in itself can cause overfitting. Also, stratified holdout does not guarantee a perfect representation of the data, therefore it is prone to produce biased error rates. One way of dealing with this is the repeated hold-out technique, which is applying the hold-out method iteratively for different chunks of the data to yield an averaged error rate (Witten, Frank & Hall, 2001, p.153).

*Cross-validation* is a form of repeated-hold out in which the data is separated into parts, referred to as folds, and each is explicitly tested on the rest of the data to find an average error rate and avoid variance (Nilsen, 82). This method of measuring prediction error is considered the standard in machine learning (Witten, Frank & Hall, 2001, p.153). When using cross-validation the convention is to use stratified tenfold cross-validation where, for all ten iterations, 9 parts of the data are used to test and the last is used for evaluation (Kohavi,1965, p.1137; Witten, Frank & Hall, 2001, p.153).

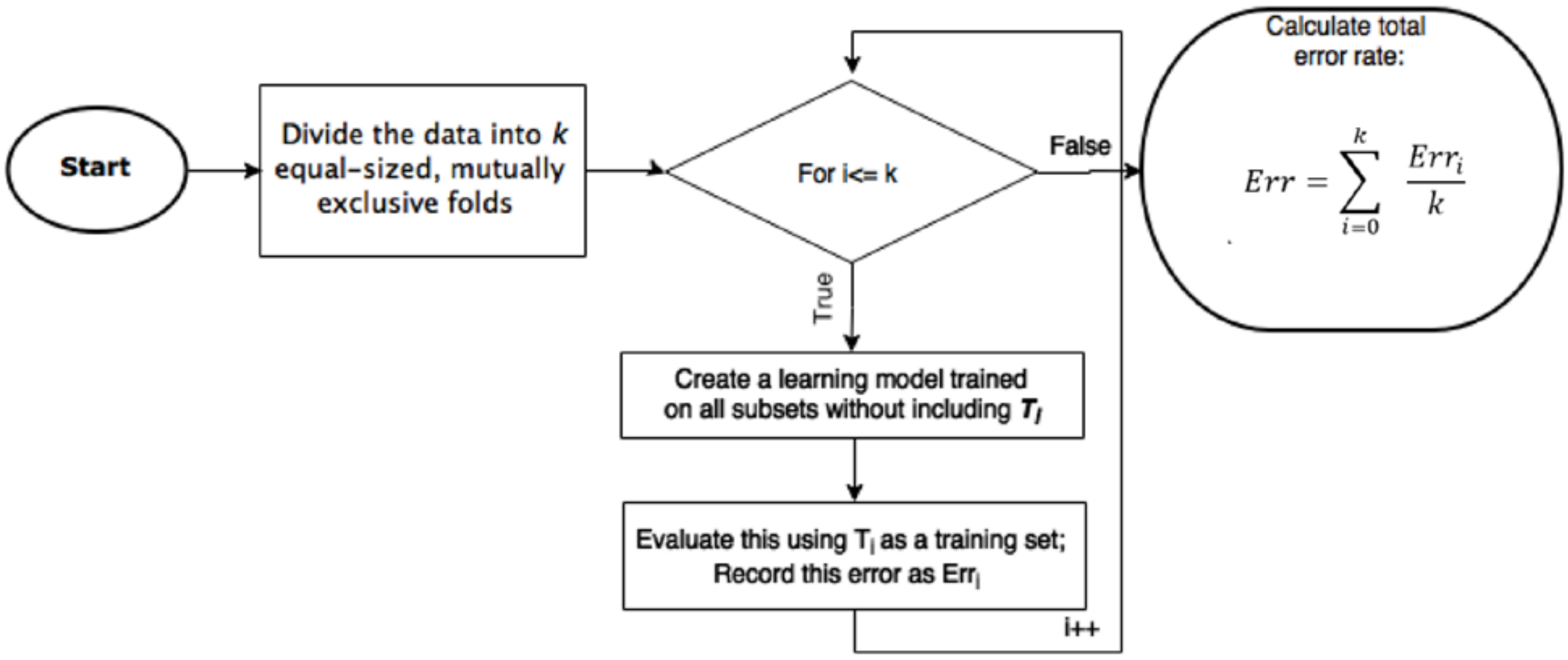

*Figure 4: K-Fold Cross Validation flowchart*

When dealing with a very restricted dataset, the *leave-one-out cross-validation* is a good choice as it can give an accurate estimate of error rate by extracting the most out of the data provided. As the name suggests, it is similar to the method outlined above, however, instead of using folds, every single instance is used as a test. Because there is no sampling involved this method yields repeatable results, but it is probably the most computationally expensive technique (Witten, Frank & Hall, 2001, p.153). To ensure the best training experience with regard to the size of the dataset, ten-fold cross validation has been implemented in all further experiments.

### *Eliminating Attributes*

Eliminating the irrelevant attributes is important for both supervised and unsupervised learning because these can create confusion, overfitting, and an increase in the computational time. It has been shown that data can become increasingly sparse when redundant attributes are provided, a phenomenon known as *the curse of dimensionality* (Mitchell, 1997, p.235). To avoid this problem, an initial feature subset selection was performed manually on the 17 attributes mentioned in section 2.1. The first attributes removed were the individual node references and the X/Y coordinates of the points because both are arbitrary values. The coordinate values are plan specific and there is a large overlap of values from the different plans (appendix A). The next attribute that was removed was the information about the teams and their polygons because these were office-unique values that belonged to 91 specific categories. These values are especially prone to overfitting because of their span and difficulty to generalise without more confidential information about the labels and the companies. Finally, the polygons that defined the individual use categories were also removed because these were directly indicative of the class value. This left each instance with 10 attributes.

These are: visual node count, point first moment, point second moment, visual mean depth, integration (based on [Tekl], [HH] and [P-value]), connectivity, visual entropy and relativized entropy.

In supervised learning there are two general ways of narrowing the concept search space, or otherwise stated – finding the minimum set of attributes that achieves the best class performance. The first method is feature subset selection and the second is dimensionality reduction. Feature subset selection is, in essence, what was done to cull the initial 17 attributes down to 10. However, when there is little or no formal understanding of the relevance of the attribute, subset selection can be performed computationally on other criteria such as consistency on class values (correlation with the classes), redundancy with other attributes (inter-correlation with other attributes) or information gain (Janecek et al., 2015). The second method of narrowing the concept search space is through dimensionality reduction. Feature subset selection can be implemented using a filter or wrapper approach. Wrappers are learner dependent methods where the data is prepared specifically to perform well on a certain learner. Filter methods are based on a statistical evaluation before any choice of learner actually takes place. This thesis will use three supervised learners: OneR, Naïve Bays and J48 to examine the attributes and test their performance on the office layouts. OneR and Naïve Bayes are implemented first because they are very basic classifiers with strong assumptions. Following the study performance of the attributes on these assumptions, a subset selection is performed, so that the more sophisticated J48 can be tested on the pure, subset and PCA dataset.

The first step to evaluating the performance of any learner is to compare it to its simplest possible scheme. The simplest learner is the ZeroR classifier, which looks at the most popular class and guesses only this for all the instances (Witten, Frank & Hall 2011, p. 448). The percentage of correctly classified instances following this method is referred to as the *baseline accuracy*. This value is important, because any classifier that performs worse that the baseline accuracy can be directly discarded. For the office dataset, the baseline accuracy is 36.92% correctly classified instances (tested with ten-fold cross-validation). Unsurprisingly, this shows that the open workspaces (G1) are the most common functions throughout the layouts, which makes sense as this configuration is most economic and flexible in terms of space usage.

### *OneR*

To establish which attribute gives the most information about the usage of the spaces, the OneR classifier was implemented. This is a simple, but powerful one-level decision tree classifier, which looks at all instances and for each attribute creates a rule that can describe the dataset best (Holte,1993). It works by looking at the least total error calculated from the frequency tables generated for each attribute. Otherwise stated, OneR looks at each possible value of an attribute, sees what the most popular class is for that attribute value and assigns all instances with that attribute value the majority class (Sayad, 2015). A frequency table for numeric data is generated by a discretisation process. This is a wrapper method that creates bins (groups) with a minimum of 6 (the classifier's default value) instances per bin.

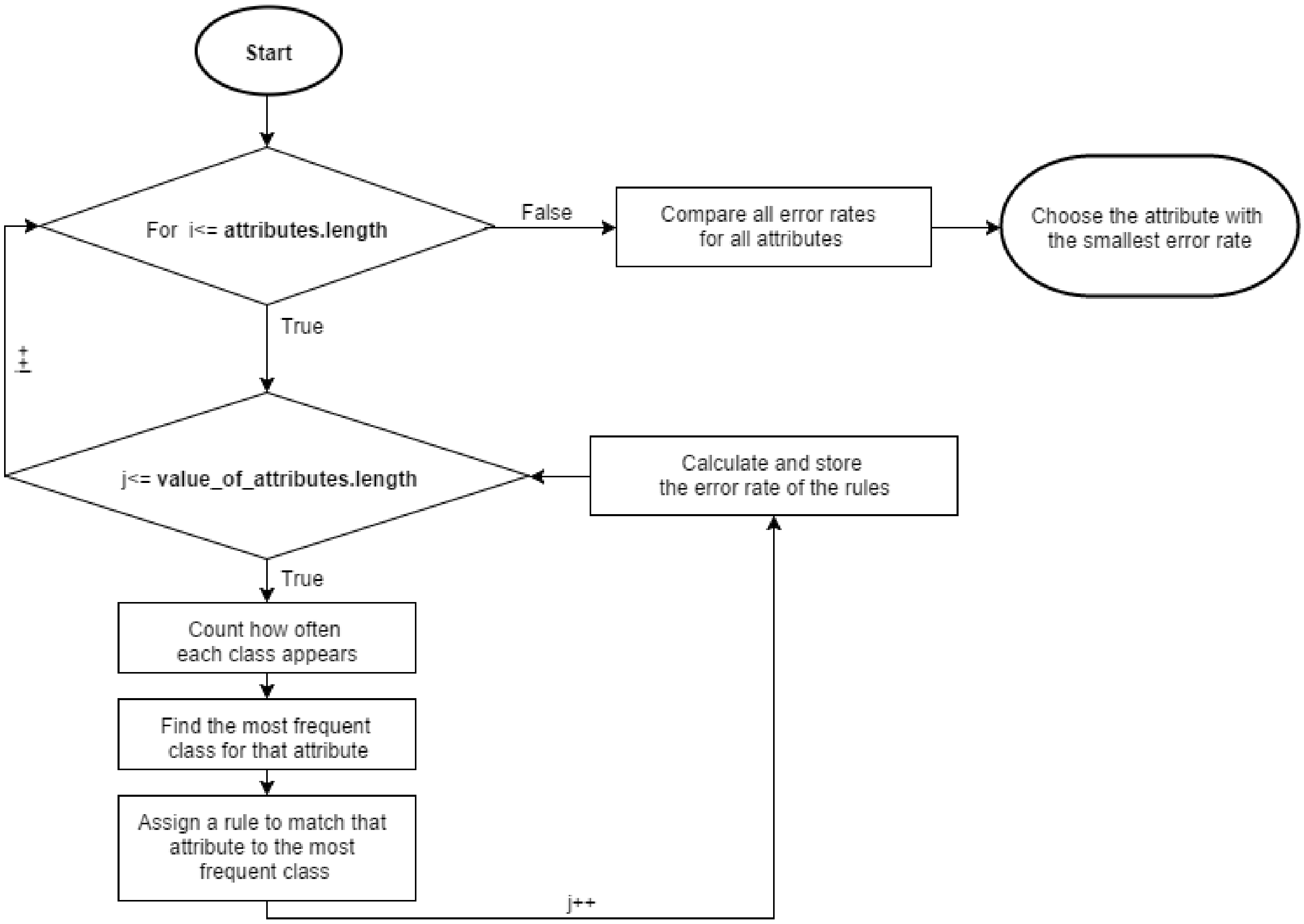

*Figure 5: OneR flowchart*

## *Naïve Bayes*

Once the relationship between the attributes and the class values had been reviewed, the question of how the attributes relate to each other was examined using Naïve Bayes. This learner has the 'naïve' assumption that all attributes are statistically independent (Brownlee, 2014; Sayad, 2015b). It predicts the class of any new instance by considering the probability of each class (given as the total occurrences for one class value divided by the total number of class values) and the probability of having that class outcome given the new instance's attribute values (refer to figure 6). For any new instance with categorical data,  It can be implemented on numerical data assuming normal distribution, however, it has been shown that this learner performs better on categorical values (Deshpande, 2015; Witten, Frank & Hall, 2011, p. 90-99.

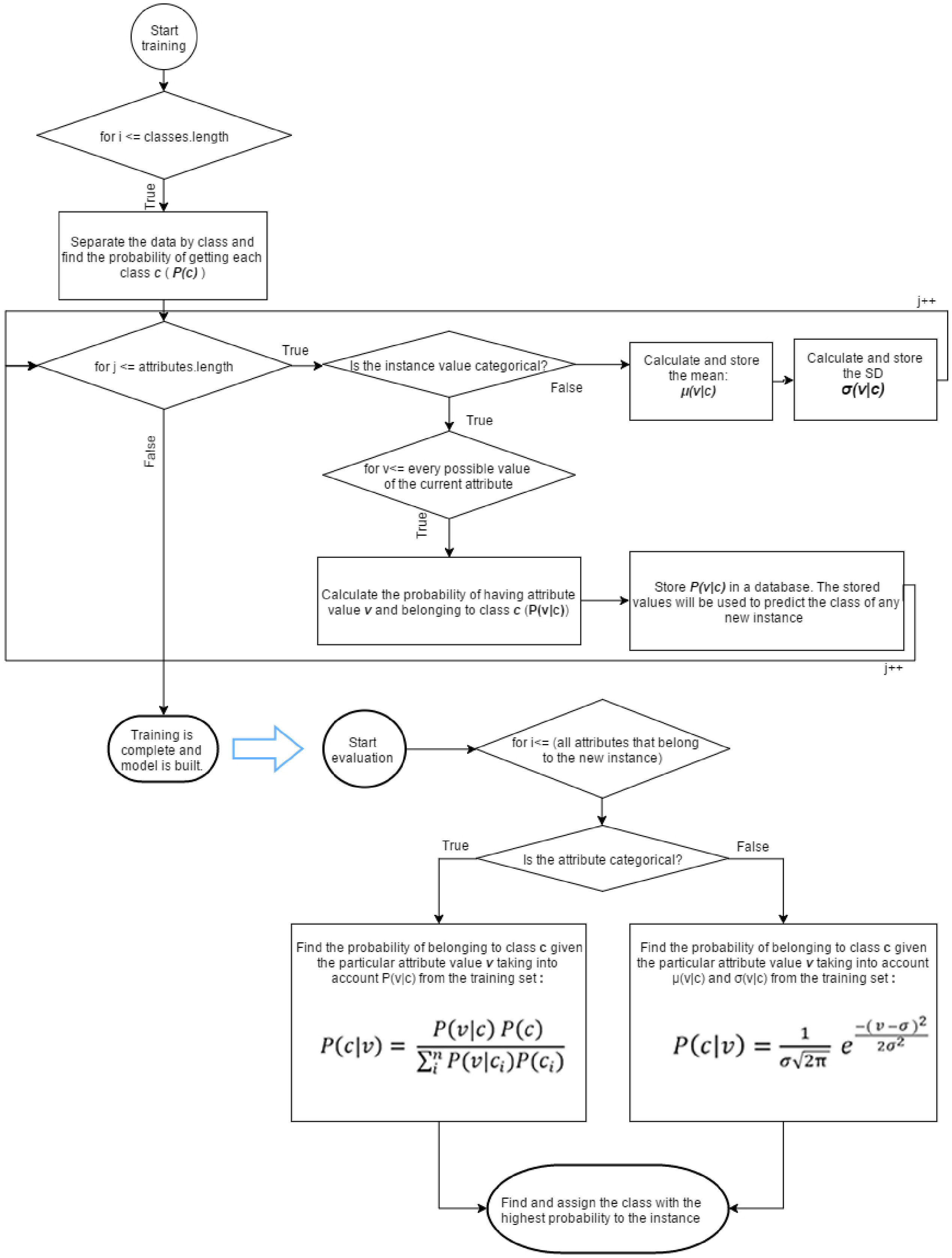

*Figure 6: Naïve Bays flowchart, formulas based on (Brownlee, 2014; Sayad, 2015b)*

The resulting subset selection is discussed in section 3.1.

### *PCA*

Dimensionality reduction was also examined as a means to reduce the attribute space used for learning. For this purpose a principal components analysis (PCA) was conducted on the full dataset. Xi provides a simple way of explaining a PCA by comparing it to visual information flattening. Conceptually it is like capturing a 3d object on a 2d image using an axonometric view. The idea is that the object is turned in such a way as to show the maximum amount of variation on either axis (Xi, 2009). What PCA does is to consider every instance as an n-dimensional vector, where n is the number of provided attributes. The data is then examined for linear attribute combinations. The line of greatest variance is found and used as a new coordinate axis, then the same is done to create an axis perpendicular to the first (Witten, Frank & Hall 2011, p. 324). This is then repeated until no more information can be gained or some stopping condition is reached, for example when 95% of the variance is achieved. The summary of the new PCA attributes and their principal component vectors is discussed in section 3.3.

## 2.2.2 Supervised Learning

### *J48 Decision Tree*

Decision trees have a general-to-specific search bias, which is useful for incorporation into a larger application and are generally more intuitive to understand. They have a hierarchical structure consisting of nodes, branches and leaves (see figure 7). Each node represents an attribute which we split on. When dealing with a continuous numeric attribute a binning process is applied to the data, so that any bin (value range) could be a potential node. Regardless of the type of attribute data, the order of the nodes is given by some purity measure to the classes. For classification there are three main metrics for purity: entropy, Gini index and the index of classification error (Teknomo, 2015). To avoid excessively large trees and overfitting, an optimization called pruning can be performed. There are two ways to do this: post pruning and pre pruning. Post pruning is when the tree is slimmed down after being completed, whilst pre-pruning occurs simultaneously to construction (Nikita and Saurabh, 2012). Pre pruning is implemented while the tree is being built. One way to do so is to set a threshold for the minimum number of instances per node (Sayad, 2015a). A node which can only be reached by a number of examples smaller than this threshold is pruned. This is useful because it is a simple way to tone the leaner, but it must be used with caution because it is very specific depending on the input dataset.

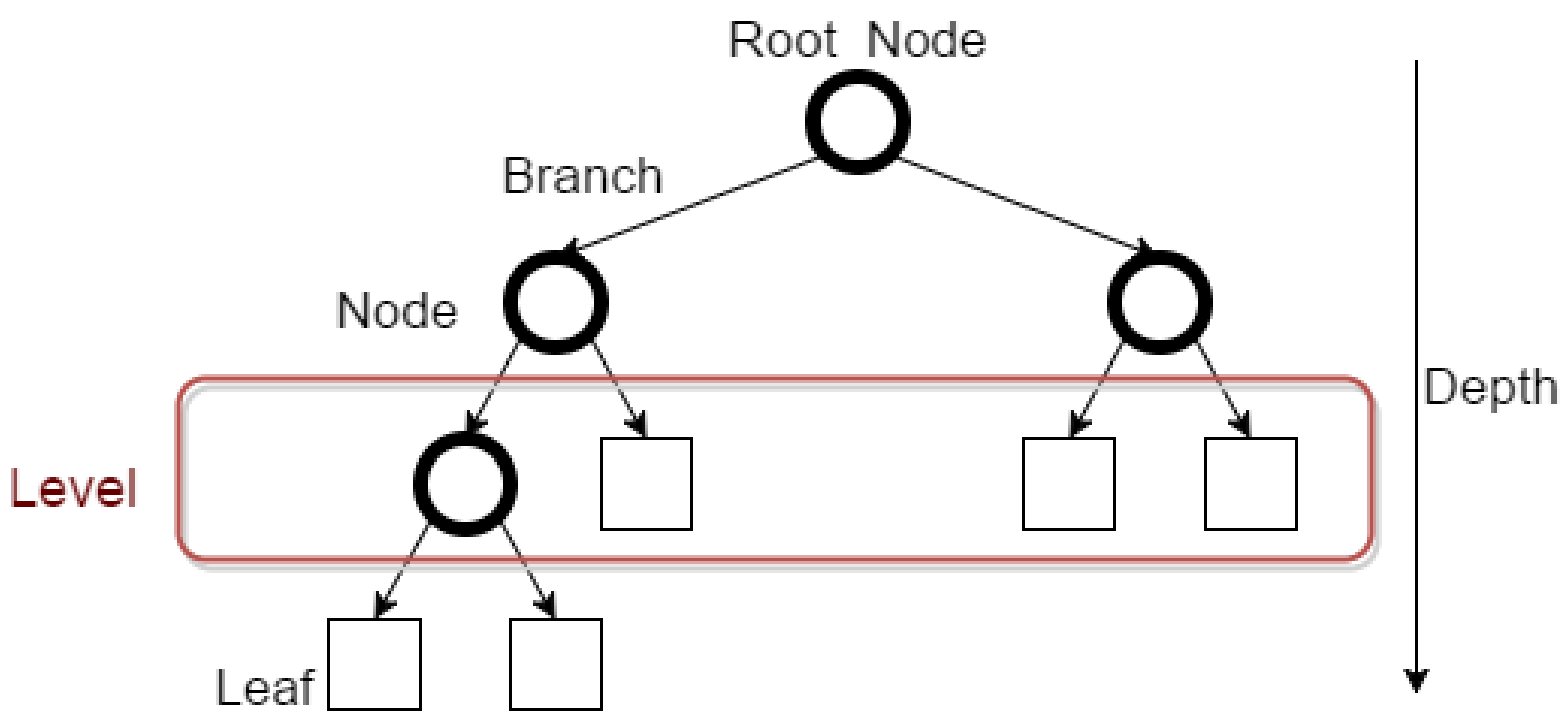

*Figure 7: Anatomy of a tree*

This study will use the java implementation of Quinlan's C4.5 revision 8, from now referred to as J48 as in WEKA. This tree uses an entropy gain criteria for splitting the nodes. The decision tree learner was tested against the office dataset with 10 attributes, 9 attributes, and the PCA vectors. This learner was chosen because C4.5 is one of the most widely implemented decision trees (Wu et al., 2007). Also, the entropy heuristic is fairly straightforward without compromising predictive strength (it must be noted that experiments with different tree learners were carried out and the difference in predictive strength was not statistically significant (<5%)). Finally, J48 is also advantageous for its reasonable run-time and heap use.

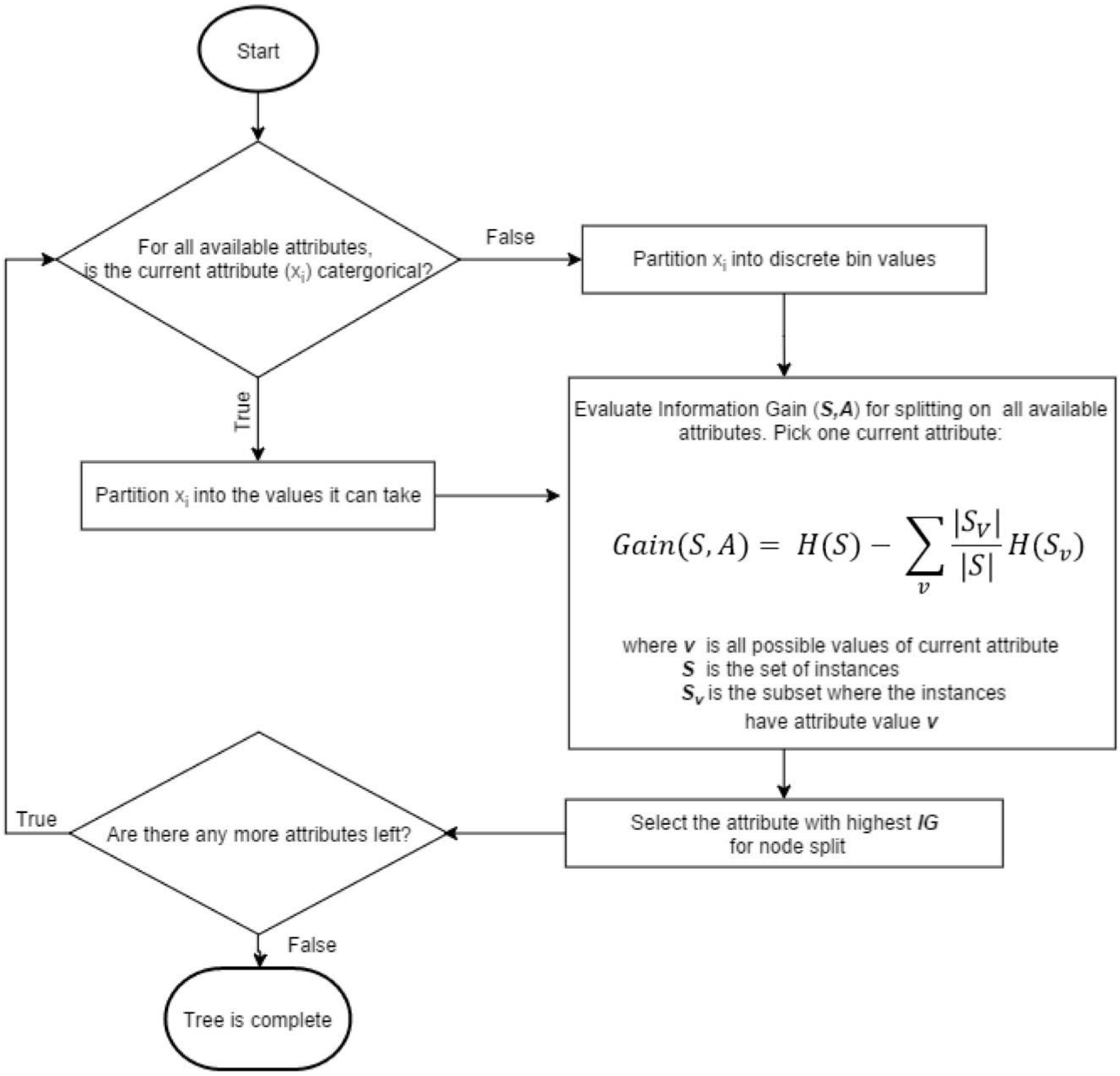

*Figure 8: J48 flowchart (Formulas based on Decision Tree 4: Information Gain, 2014)*

The learner is implement once with a three-fold, reduced error pruning and once with error-based pruning (figures 10 and 11).Reduced error pruning (Quinlan, 1999) is done by evaluation on a part of the training set, referred to as the validation set. Error-based pruning does not require a validation set but instead works with a confidence factor (Crane, 2015). For both implementations, the minimum number of instances per leaf is specified at 50. This is chosen to roughly estimate a small space of 10m$^2$ (note, an instance is the centre node of a 0,45m x 0,45m square).

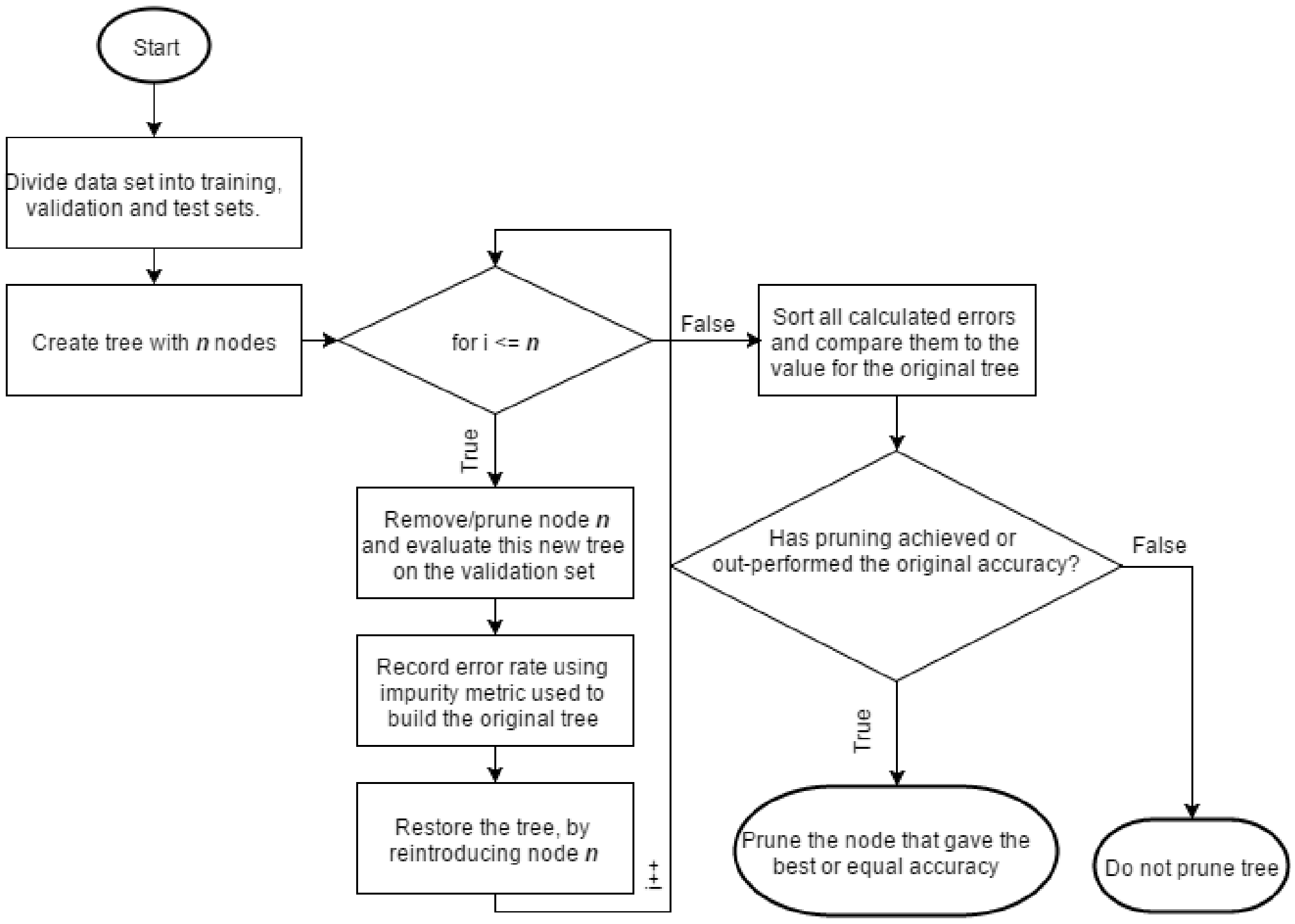

*Figure 9: Reduced Error Pruning Tree*

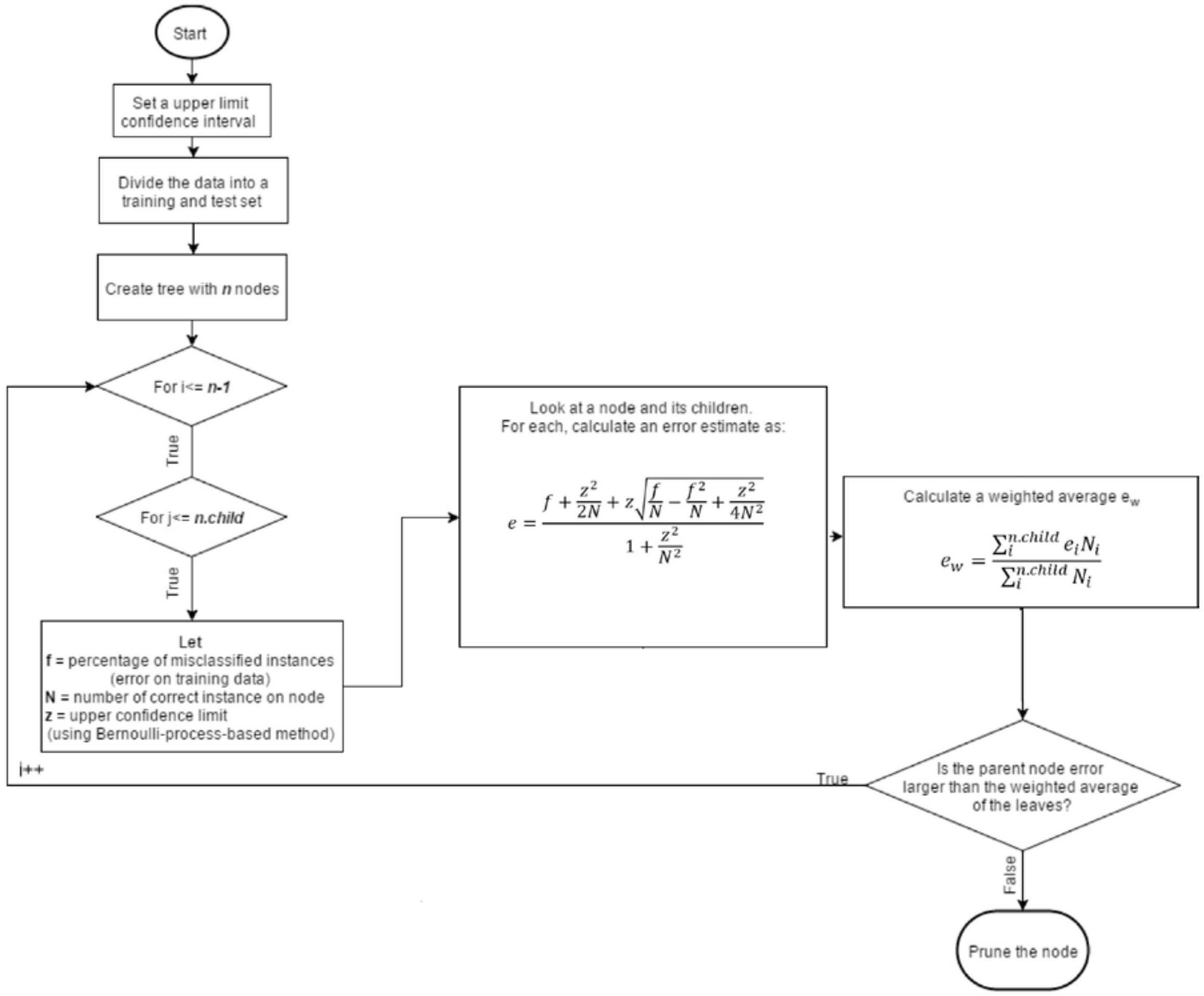

*Figure 10: Error-Based Pruning Tree (Formula based on Crane, 2015)*

### 2.2.3 Unsupervised Learning

Two unsupervised learners, K-Means and Self-Organising Maps, are tested to see if the inherent data structure of the VGA measurements is able to account for an adequate classes-to-clusters evaluation. These have been included in the results section, but will not be discussed in detail here because they were less illuminating and serve only a secondary purpose in the investigation. Nonetheless, an overview of these can be found in Appendix D.

# Results

## 3.1 Eliminating Attributes

*OneR*

| Rank | Attribute | Correctly Classified |
|---|---|---|
| 1 | Visual mean depth | 67.896 % |
| 2 | Integration [P-value] | 67.724 % |
| 3 | Integration [Tekl] | 67.597 % |
| 4 | Integration [HH] | 67.521 % |
| 5 | Visual Entropy | 59.923 % |
| 6 | Visual Relativized Entropy | 59.855 % |
| 7 | Connectivity | 49.190 % |
| 8 | Point First Moment | 43.886 % |
| 9 | Point Second Moment: | 42.813 % |
| 10 | Visual Node Count: | 42.212 % |

*Table 3: Ranking of attribute relevance based on the OneR wrapper (OneRAttributeEval in WEKA)*

OneR tests show that the most descriptive attribute with the best, class-wide performance for the office dataset is visual mean depth (refer to table 3). It performed best, however not significantly better with regard to its normalised version (integration Tekl/P-value/HH). The next attributes by virtue of predictive strength, although with significantly less accuracy, are visual entropy and visual relativized entropy. These indicate the likelihood of taking a turn in the next step, or otherwise stated, the perceived orderliness of the layout, could determine the usage class with at least 59.9% accuracy.

A possible explanation for this is that interiors of offices tend to be organized in a similar way. For example, open-plan spaces, which accommodate the majority of employees, would tend to be in the vicinity of other open functions, such as primary circulation, print/copy facilities and the reception. Functions like meeting rooms would be in more undisturbed areas, away from kitchen and tea areas, and storage would be expected to be in secluded corners. It could be the case that efficiency of movement (depth), rather than visual perception of space is more important in the office context because the people in the space would more often be occupants, who are familiar with the space, as opposed to visitors, who would need to find their way around. Finally, one could expect the organisation of the interior to always be subject to the building and site-specific. This could be another reason why entropy values rank second in predictive strength to mean depth and its derivatives.

As for the remaining four attributes, these make a correct prediction for little less than half the instances, but manage to predict above the baseline accuracy. The ranking shows that the number of nodes seen, as given by connectivity, is more indicative of the class than point first moment, which gives the distances visible nodes in the system. The latter is, in the general case, more informative because it provides insight into how far it can be seen into the rest of the layout, but also how compact the perimeter of its isovist is (see section 2.1.2). Nonetheless, because the open work spaces, which are the majority class in this building type are characterised by a high connectivity, it outperforms point first moment. Finally, it can be noted that the size of a layout, given by the node count, is least indicative of what usage types a certain node will belong to. This can be expected because any office, large or small, will have approximately the same functional spaces for this type of occupation.

#### *Naïve Bayes*

For the office dataset, using numeric attributes, the Naïve Bayes classifier performed below the baseline. The classifier's performance was slightly improved by applying WEKA's unsupervised nominalisation filters, however the performance remained relatively poor. Table 4 shows that binning improves percentage performance and class performance for all classes. After various trials, in which attributes were removed, it was found that even the best performance of Naïve Bayes, a subset of 3 instances, was incorrect in more than half of its prediction and with ill class coverage (see table 4/ figure 11). The reason for this is that the underlying assumption of this learner does not match the structure of the data. The attributes are mostly co-dependent because, of the 9 values other than visual mean depth, 5 are directly dependent on it: integration ([Tekl]/[HH]/[P-Value]), entropy and relativized entropy; and 3 attributes are based on connectivity (point first moment, point second moment and, of course, connectivity itself). The only truly independent attribute in the set is the node count, which is the basis for all visibility calculations.

| Attributes used | Number of attributes | No Nominalisation | Equal (10 Bins) |
|---|---|---|---|
| Complete Attribute set | 10 | 20.7375 % | 38.3962 % |
| Visual Node Count<br>Point First Moment<br>Visual Mean Depth<br>**Visual Integration [Tekl]**<br>Visual Entropy<br>Visual Relativised Entropy<br>Point Second Moment<br>Connectivity | **8** | **23.3005 %** | **41.1973 %** |
| Visual Node Count<br>Point First Moment<br>Visual Mean Depth<br>**Visual Integration [HH]**<br>Visual Entropy<br>Visual Relativised Entropy<br>Point Second Moment<br>Connectivity | 8 | 22.0284 % | 40.9116 % |
| Visual Node Count<br>Point First Moment<br>Visual Mean Depth<br>**Visual Integration [P-Value]**<br>Visual Entropy<br>Visual Relativised Entropy<br>Point Second Moment<br>Connectivity | 8 | 21.0624 % | 41.0414 % |
| Visual node count<br>Visual mean depth<br>Visual Relatives Entropy<br>Connectivity | 4 | 35.95 % | 48.1079 % |

*Table 4: Performance of the Naïve Bayes learner with different nominalisation procedures applied to the numeric attributes*

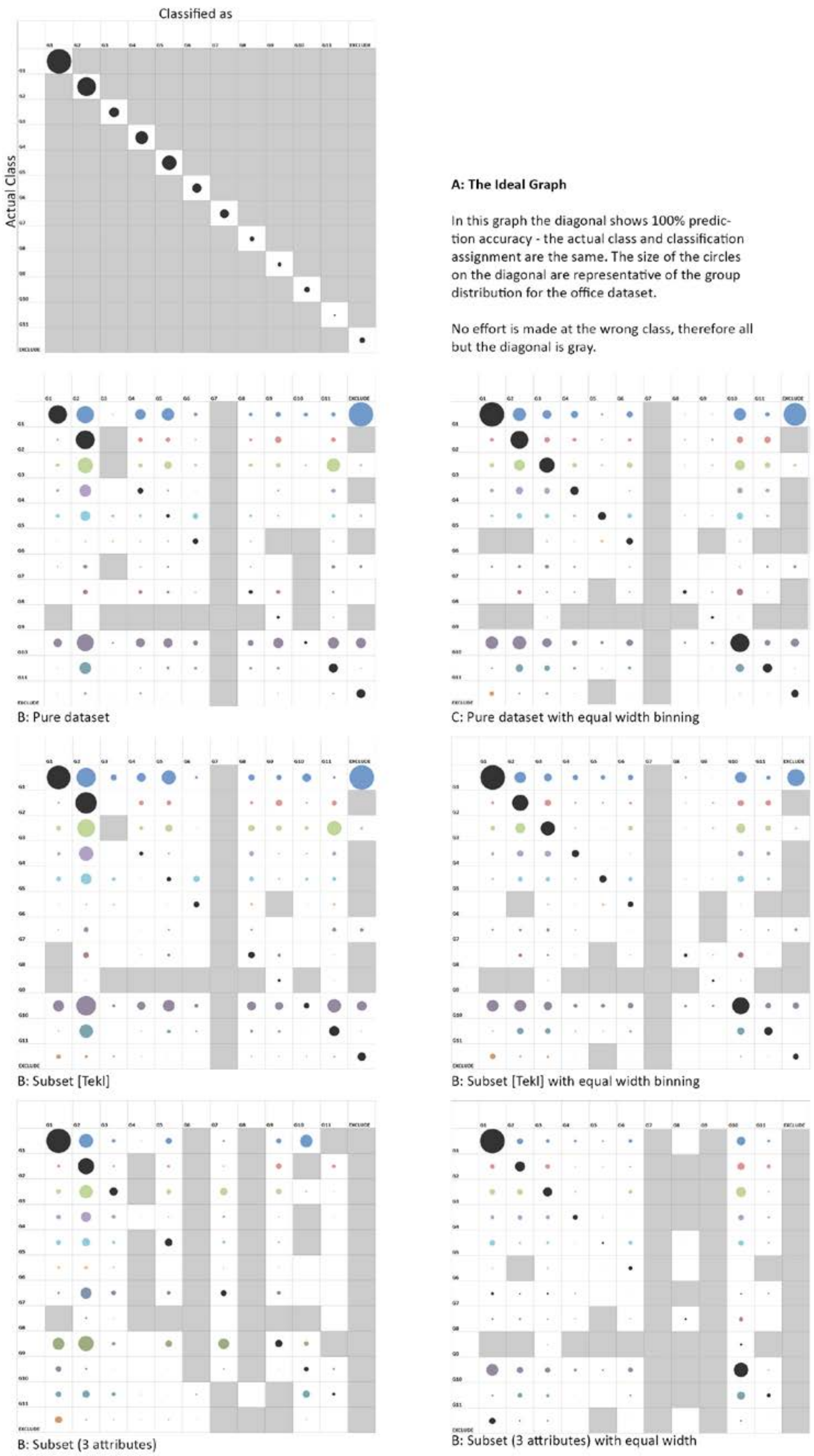

*Figure 11: : Confusion matrixes as heat maps. It can be seen that the most confusing, for which little to no effort of guessing was made is G7 (Print and Copy Facilities), this is likely because of the category considers both rooms and single machines. The most distinct category with least wrong guesses is G9 (Extra Facilities).*

Taking into account the dependencies, a repeated test with a smaller number of attributes was done to validate that the interdependencies of the data were having a negative effect on the data. This showed that of the integration values, Integration Tekl is the most independent. These are explored more closely in the following section.

## 3.2 Detailed Integration Study

This section examines the different integration values in more detail, as a follow up to the results obtained through the tests with OneR and Naïve Bays. From the supervised tests, two main conclusions on integration could be deduced. Firstly, of all the integration values, P-Value appears to be slightly more descriptive on its own for classification with a predictive strength of 67.72% (compared to 67.60% (Tekl) and 67.52 % (HH)). Secondly, the Naïve Bays showed that, of the integrations, the Tekl measure was least dependent on the other attributes provided by yielding the highest prediction (up to 41.20 %). A more detailed study on integration has been done because the literature of these presents different ways of calculating a normalised mean depth and it is focused exclusively on urban areas (Teklenburg et al., 1993; Krüger, 1989; Turner, 2001/2008; Hillier, 2007; Al-Sayed, Alisdair and Hillier,2013). Taking into account the size of the data, the measurement's performance could be evaluated in a wide span of cases with varying sizes.

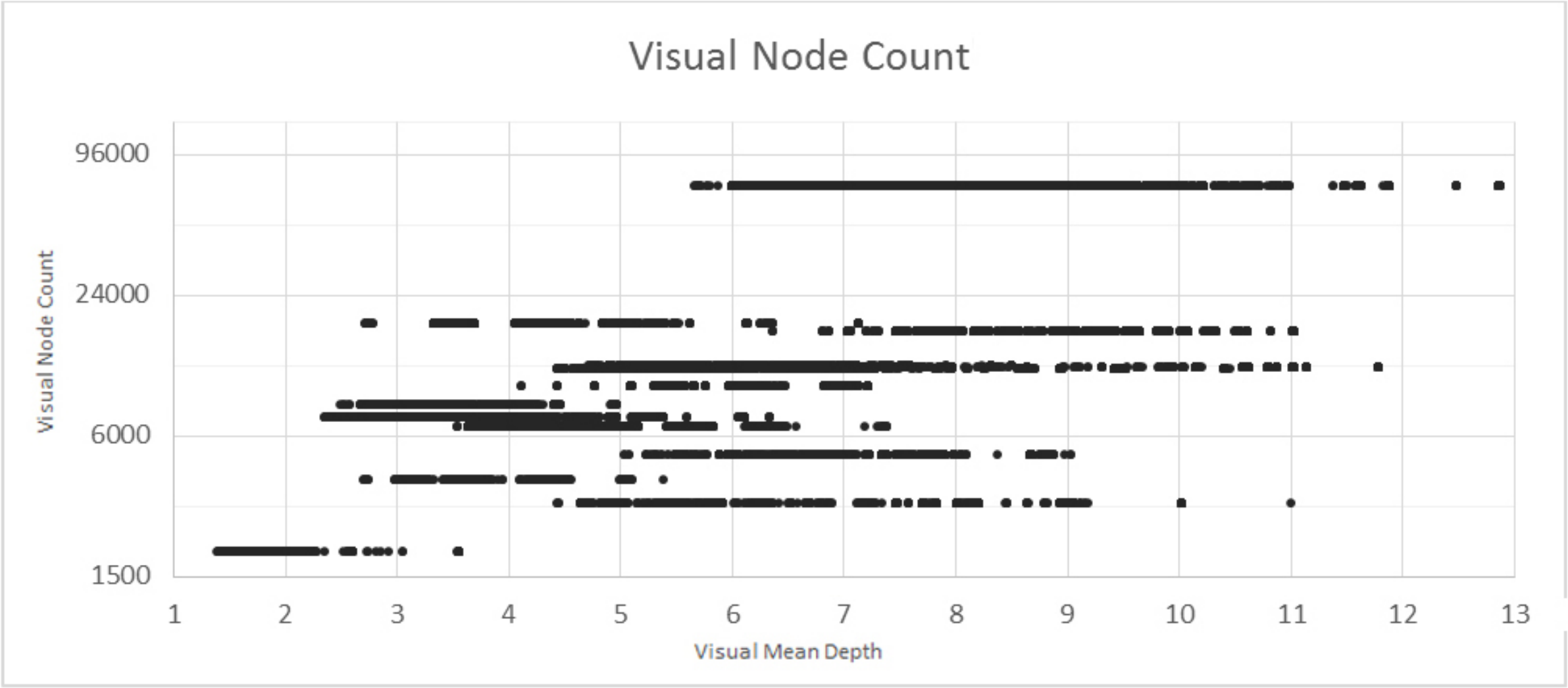

*Figure 12: A graph of node count against visual mean depth*

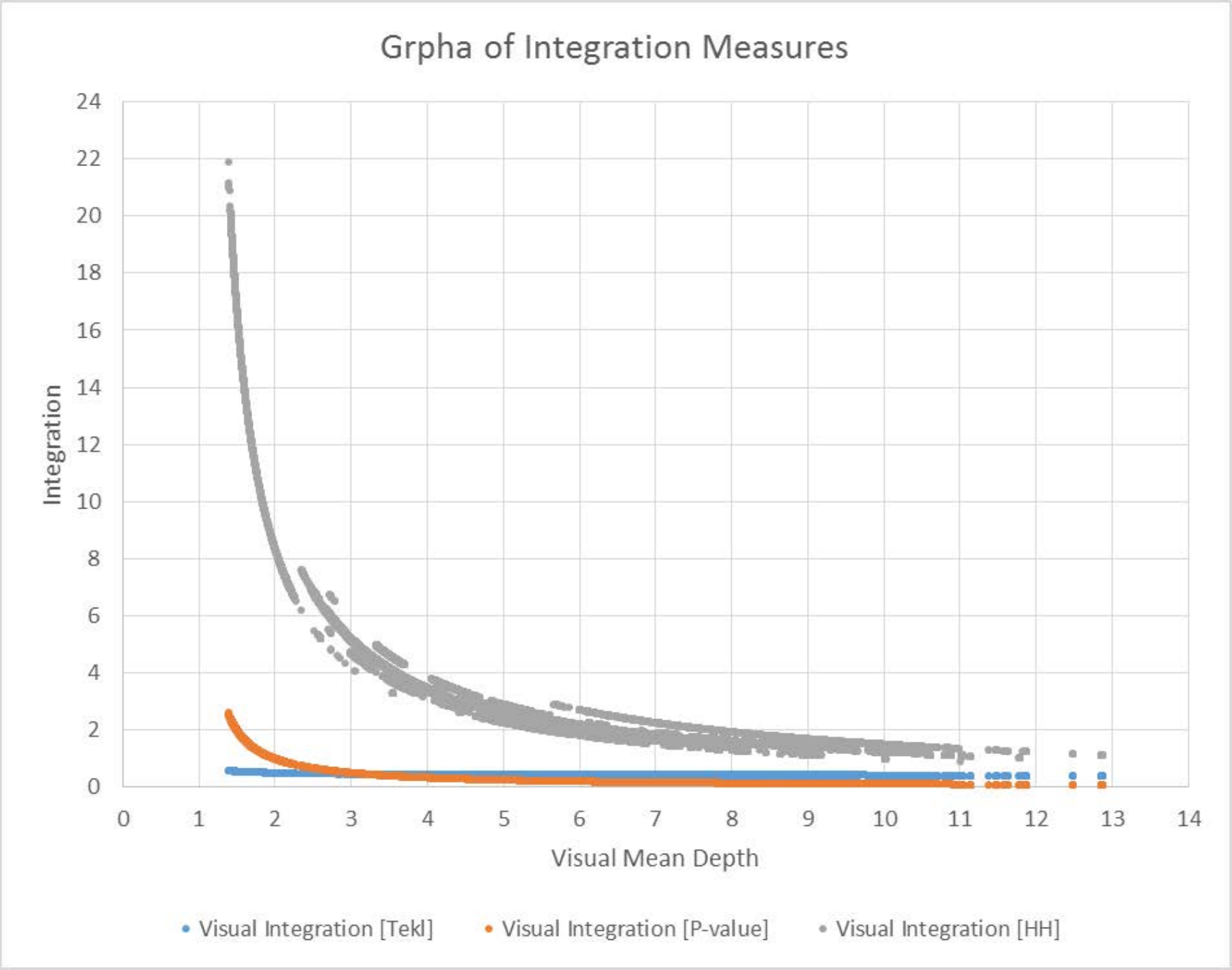

*Figure 13: A graph of the different integration values against visual mean depth*

| | Mean Depth | Integration | | |
|---|---|---|---|---|
| | | **[P-Value]** | **[Tekl]** | **[HH]** |
| **Minimum** | 1.381 | 0.084 | 0.389 | 0.902 |
| **Maximum** | 12.873 | 2.609 | 0.573 | 21.884 |
| **Mean** | 6.54 | 0.225 | 0.43 | 2.512 |
| **Standard Deviation** | 2.075 | 0.166 | 0.017 | 1.446 |
| **Variance** | 4.3056 | 0,0276 | 0.0003 | 2,0909 |

*Table 5: Statistical characteristics of integration values*

Figure 13 shows the relationship of visual mean depth to integration P-value, Tekl and HH for 14 case studies with a total of 36 floor plans. From figure 12 and figure 13, it can be observed that as mean depth increases, so does the similarity of the results of the different integrations measures. As mentioned previously in section 2.1.2, the purpose of all three integration measurements from space syntax is to eliminate the layout size bias, so as to allow for a better comparison of mean depth from different sized layouts. It can be observed that although there is only one case study for mean depth between 1 and 2 there is a great variance in integration values for integration HH and P-value. It is also worth noting that the integration HH curve is least consistent, and diverges multiple times between mean depth 2.5 and 11, where multiple case studies are present. Integration Tekl achieves the highest normalisation consistency, as it has the lowest variance for the different mean depth values (table 5). This measure shows the least variance because it is calculated using only the mean depth of the given layout against the mean depth of an equivalently sized bipartite graph (see figure 14, far right). In the other two cases the normalisation is achieved by means of a comparison between the asymmetry of the root node of some other standard graph: the diamond root graph for HH and corner-of-a-grid graph for P-value, which includes approximately the same number of nodes.

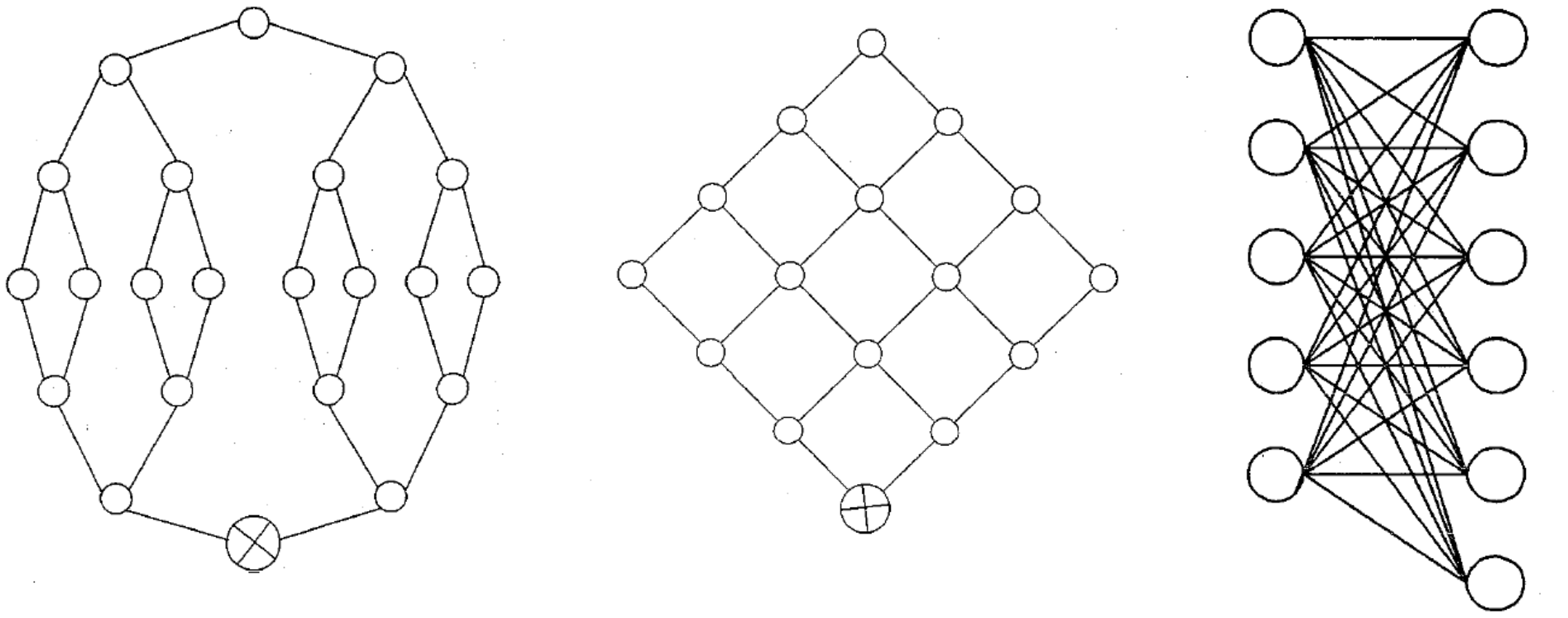

*Figure 14 (Teklenburg, 1993): Diamond root Graph, corner-of-a-grid graph and bipartite graph used for normalisation of Integration HH, P-value and Tekl, respectively. The root node for the asymmetry calculation for the first two is marked with an 'x'*

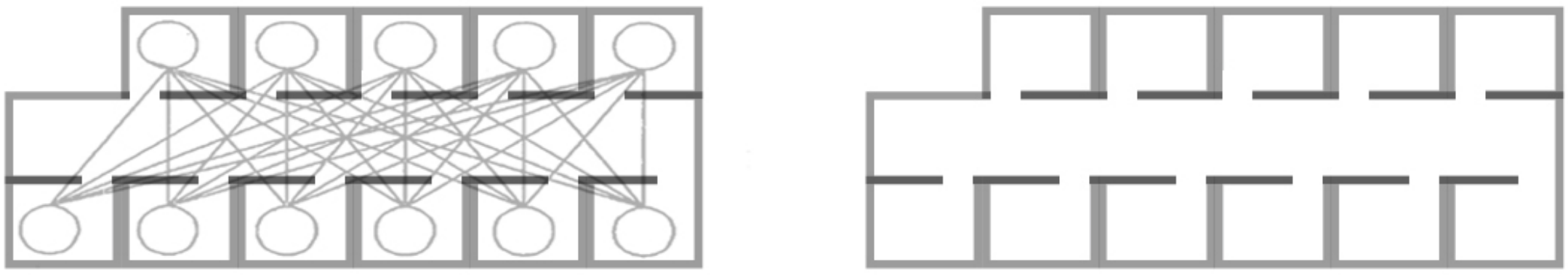

*Figure 15: Architectural translation of the bipartite graph for an interior space. The other two have not been recreated as interior situations because these have not been found to have a single, unique translation.*

Perhaps the asymmetry calculation itself is not as problematic for consistency as is the nature of the comparison graphs. The bipartite graph has the simplest growing rule: if the nodes can be divided equally then both sides are equal, if not then then a graph such as the one in figure 14 (far right) is created (Teklenburg et al., 1993). This makes it possible to always have the same number of nodes in the bipartite graph as in the layout that is being normalised, regardless of size (which is not always the case for other two standard graphs, especially the Diamond root graph used in Integration HH). The large differences between the Integration values, especially for smaller floor plans with less node counts, is because the different methods of growing flexibility of their respective standard graphs.

The graph used for Integration HH grows exponentially (in powers of 2) from each endnote (top and bottom) to the middle. By comparison, the P-value standard graph is grown as an arithmetic sequence, therefore the increase is less and the scope of the measure’s values in it is smaller (figure 13).

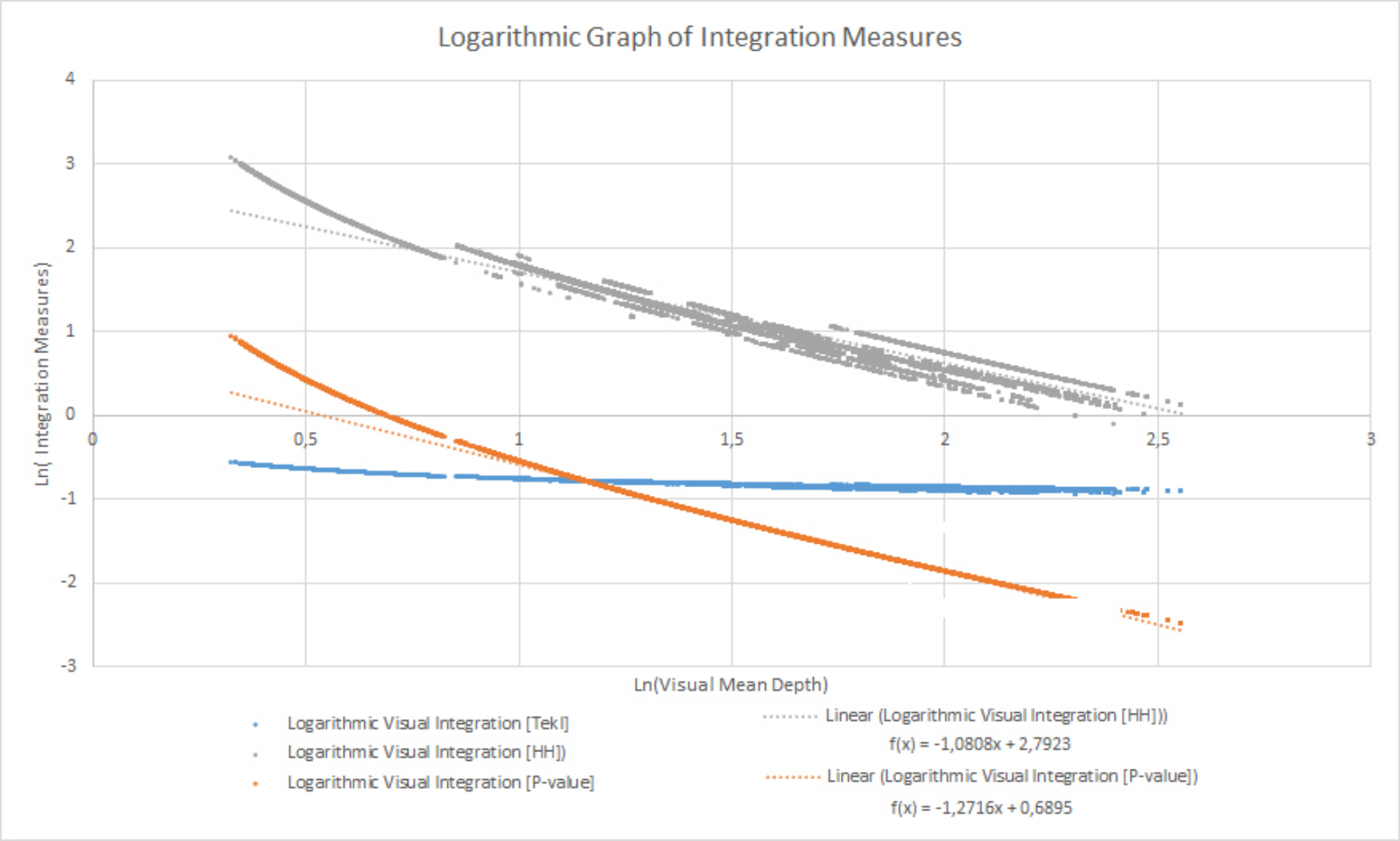

*Figure 16: a logarithmic graph of integration values versus visual mean depth*

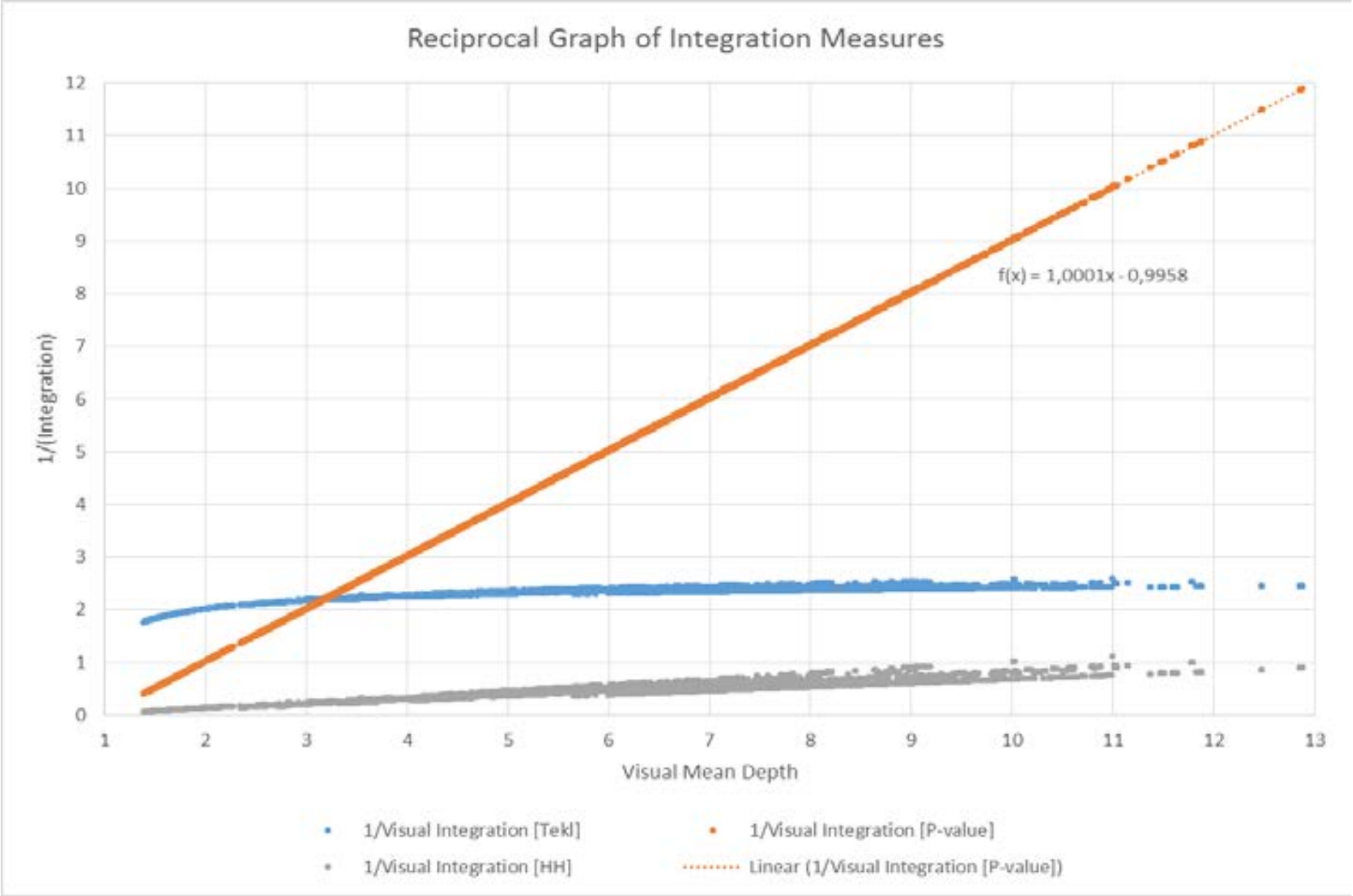

*Figure 17: a graph of reciprocal integration values against the visual mean depth*

From figure 13, it seems that the integration values have either an exponential or a reciprocal dependence on visual means depth. A closer examination of this was done in figure 16 and 17. The former shows a plot of the natural logarithms of the integration values against the natural logarithm of visual mean depth. It shows that their relationship is not strictly exponential because the integration values do not appear linear. Figure 17 shows the reciprocal of the integration values plotted against visual mean depth. It shows that the P-value measure and the HH are close to being a straight line. The P-value especially demonstrates a very predictable linear dependency. As for integration Tekl, this is not as linear as expected based on its formula (see Teklenburg et al, 1993 p.354).

In conclusion, this section showed that of the HH value is not a reliable normalisation technique for interiors, because its curve diverges as different sized layouts are introduced. The study also showed that Tekl has the smallest span and variance, whilst the reciprocal of P-value integration is most predictably related to visual mean depth.

## 3.3 Principal Components Analysis

| 4 V1 | V2 | V3 | V4 | V5 | Attribute: |
|---|---|---|---|---|---|
| -0.1669 | -0.3414 | -0.7662 | 0.2311 | -0.299 | Visual Node Count |
| 0.0542 | -0.5529 | 0.1749 | -0.0361 | 0.0525 | Point First Moment |
| -0.4038 | -0.0294 | -0.3334 | 0.1404 | 0.3499 | Visual Mean Depth |
| 0.4344 | -0.0572 | -0.189 | 0.0796 | 0.0199 | Visual Integration [Tekl] |
| -0.3698 | -0.0613 | -0.1262 | -0.553 | 0.5456 | Visual Entropy |
| -0.3523 | -0.0058 | 0.3151 | 0.7489 | 0.2635 | Visual Relativised Entropy |
| 0.0329 | -0.5351 | 0.1785 | -0.0908 | 0.0877 | Point Second Moment |
| 0.0797 | -0.5301 | 0.1662 | 0.0351 | 0.0126 | Connectivity |
| 0.4103 | 0.0498 | -0.1493 | 0.1573 | 0.5282 | Visual Integration [P-value] |
| 0.4252 | 0.0016 | -0.1961 | 0.1356 | 0.362 | Visual Integration [HH] |

*Table 6: New PCA base vectors and their relation the original attributes*

A principal components analysis of the attributes created 5 vectors, together capture 88.14 % of the variance of the full set. The new PCA attributes were tested with Naïve Bayes to confirm that these are independent. The performance of the learner with the treated attributes was indeed doubled, with 44.38% prediction accuracy without nominalisation and 46.67% with ten, equal-width binning. This shows that the PCA attribute achieves low inter-correlation of the attributes whilst retaining most of the variance.

## 3.4 Supervised and Unsupervised Learning

### 3.4.1 J48 Decision Tree

| | Reduced Error Pruning (REP) | | | Error Based Pruning (EBP) | | |
|---|---|---|---|---|---|---|
| | **Raw** | **Subset** | **PCA** | **Raw** | **Subset** | **PCA** |
| **Prediction Accuracy** | 77.7749 % | 77.7006 % | 76.505 % | 79.4654 % | 79.3361% | 78.5419 % |
| **Tree Size** | 1207 | 1227 | 1313 | 1665 | 1677 | 1651 |
| **Leaves** | 604 | 614 | 657 | 833 | 839 | 826 |

*Table 7: Summary of prediction results and tree size from different J48 experiments*

The results of the J48 learner show that a prediction accuracy of up to 79.4654 % can be achieved by using the complete attribute set with the EBP. It can be observed that for both pruning scenarios, the complete dataset outperformed the subset by a very small margin and the PCA performed slightly weaker. This is likely due to the small loss in variance, which occurred with dimensionality reduction.

In general it can be observed that EBP gave slightly better predictions than REP - up to 2% in the case of the PCA attribute set. This is likely because in the actual training set for REP is smaller, a part of it is kept for validation (the actual pruning). It can also be seen that there is a trade-off between accuracy and tree size. The simpler REP generated on average smaller trees, with less leaves, which could be again because was given a smaller collection of data to build a model on.

### 3.4.2 K-Means and Self-Organising Maps

| **K-Means** | **Prediction Accuracy** | | |
|---|---|---|---|
| | **Raw** | **Subset** | **PCA** |
| **Complete dataset** | 22.77% | 21.42% | 22.76% |
| **Single largest layout** | 26.32 % | 26.16% | / |

*Table 8: Summary of K-Means experiments*

| **SOM** | **Prediction Accuracy** | | |
|---|---|---|---|
| | **Raw** | **Subset** | **PCA** |
| **Complete dataset** | 22.18 % | 23.45 % | 24.27 % |
| **Single largest layout** | 33.70 % | 32.06 % | 36.89 % |

*Table 9: Summary of SOM experiments*

Overall it can be seen in table 9 that the SOM performed better than K-Means clustering, but classes to clusters evaluation is weak for both. It could be that the strongest inherent structure in the dataset is the sizes of the layouts, therefore it when given the whole dataset the clusters are formed in layout size-ranges. This claim is would also explain why the PCA attribute set did so well in both cases. To test this hypothesis the clusters were tested also on the single largest layout with 70 354 instances, which had an example of each class. This implementation did field better results for all cases and was very close to the baseline accuracy (36.92% as discussed in section 2.2.1) by a very small margin when used with the PCA (see appendix D), therefore it could be argued that large scale clustering of layouts is problematic because of lack of good ways to normalise their measures.

# Conclusion

This thesis examined patterns of spatial affordance in 14 case studies with 36 floor plans of office buildings in the UK using three supervised and two unsupervised learning techniques. This has been done to inform the search spaces for new layout designs. The background research showed that although it is possible to build tools for semi- and automated layout generation, these have not been well accepted because of their interaction quality and runtime. It was suggested that such systems could be improved by having a better understanding of the implicit connection between geometry and usage. This dissertation contributes to the better formal understanding of the relationship between the two, by looking at measurements from space syntax and usage in office layouts.

The methodology implemented is intended to be generally applicable beyond the office layout type. The attributes were defined as values taken from the VGA analysis and have been thoroughly explained. Establishing this frame work allows for testing beyond the office dataset with comparable and replicable results. Testing with the WEKA tool's library also allows room for future development and adaptation, as the source code of every learner is openly available.

The dissertation evaluated the geometry-to-function relationship of 176 395 instances, with a complete set of 10 attributes, a subset of 8 and a further subset of 5 principal component vectors. The two attributes culled from the subset – Integration [HH] and integration [P-value] – were removed after a closer investigation of the attributes using the classifiers OneR and Naïve Bayes, followed by a more detailed investigation of the three integration measurements.

Two hypothesis were put forward in this work. The primary one was that there is a link between geometry and usage, therefore supervised learners can be trained to identify use based on information from a VGA. This claim was tested against a baseline accuracy measure of 36.9% correct prediction. It was shown that even using the simplest learner, OneR, rules can be made based only on visual mean depth or the different integration measures that can predict accurately for up to 67.9% of new examples. OneR's results also showed that physical measures are a better predictor of usage compared to the purely visual measures given by visual entropy, connectivity and other values derived from them. This is true for the office dataset, but it does raise the question of whether this would hold true for other layout types. It could be the case that, for public buildings such as commerce centres or museums, visual perception is more important for usage than physical depth because the main users of the space will be visitors, rather than occupants. The visual distance and orderliness of a space could be more indicative of usage when ease of navigation is more important than speed or efficiency. Finally, the first hypothesis was further confirmed after implementing the more sophisticated J48 decision tree learner, which yielded up to 79.6% correct predictions, which is almost double the baseline accuracy.

The second hypothesis presented in this dissertation was that usage can be reverse-engineered based on VGA measurements to achieve a strong cluster-to-class evaluation. This claim was tested using K-Means clustering and Self-Organising Maps. It was not confirmed because even the best case – the SOM tested with only a single layout – achieved only up to 36.9% accuracy. The hypothesis was not confirmed possibly because the VGA did not provide enough usage-relevant information for the office dataset. It is also possible that the social aspect of a space is more strongly related to use than implicit geometry.

In conclusion, the results of this research are threefold. It outlines a methodology that can be reused and expanded for future investigations. Several classifier models were built in WEKA, which could be optimized for implementation within a larger practical tool for layout design. Most importantly, it showed that a learner can be trained to evaluate use based on only on a VGA and suggests further research directions for improved measurements, which could allow for use to be reverse-engineered without human supervision.

Word count: 10 168

# Appendix A

## Group 11: Secondary Circulation

After a closer examination it was established that the missing values in the data set, represented circulation spaces not integral to the offices (Figure A, B). Below are the polygons that define the use of the spaces and the corresponding layout with the visibility graph overplayed. The values marked as having missing a usage tag are the spaces that lead to the common staircase.

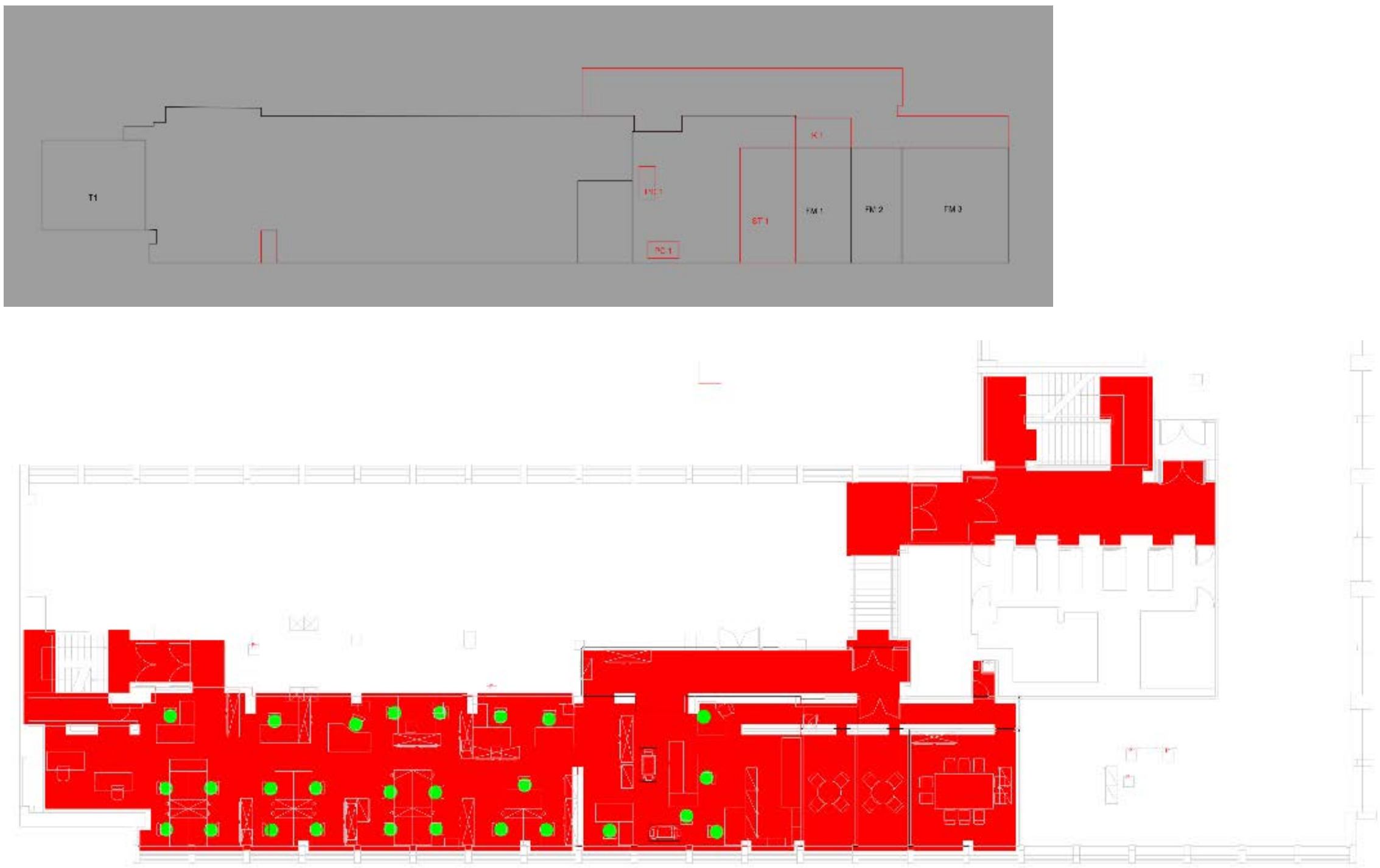

*Figure A1: Missing values can be seen in the layout in upper left corner (Case study 6, floor plan LG)*

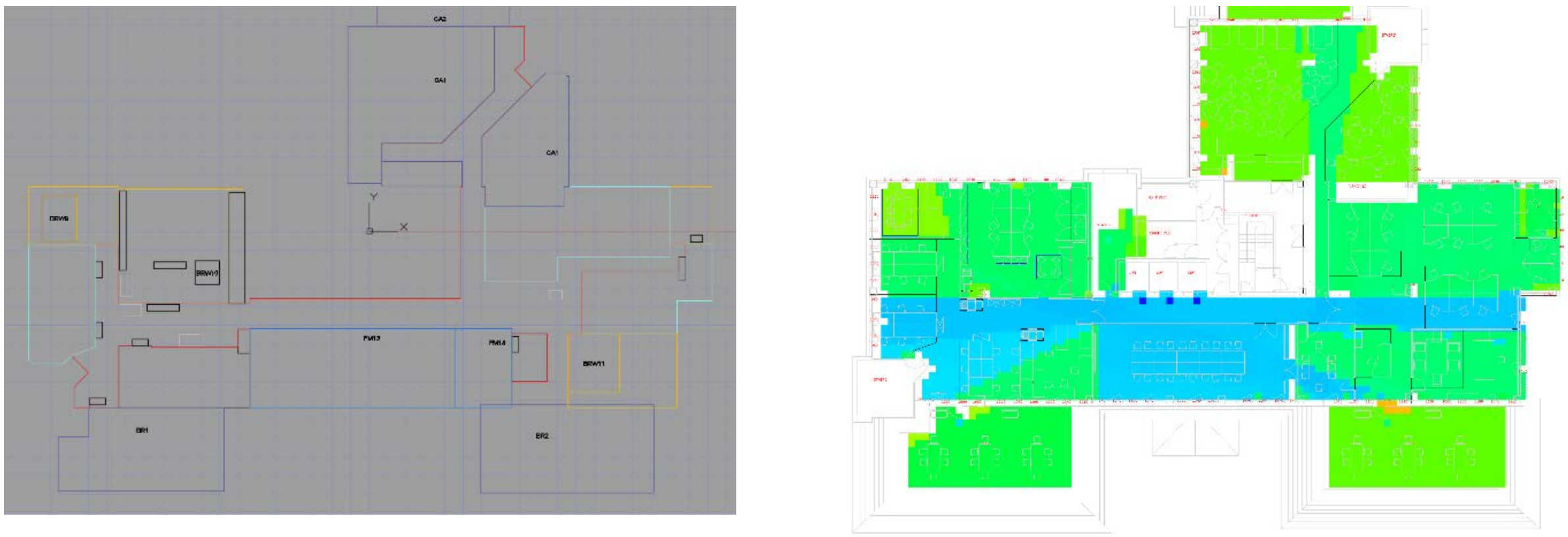

*Figure A2: Missing values can be seen in the middle of the layout (Case study 9, floor plan GR)*

## Grouping Details and SpaceLab Labels

| SpaceLab labels | Number of Instances | Percent of label to all others in Class | Class | Total instances in class | Percent of Class to all others |
|---|---|---|---|---|---|
| WRKSP-OPN | 61730 | 0,947651 | **G1:Open Workspaces**<br>Open plan desk area, alternative open work spots near main workspace and other alternative workspaces | 65140 | 0,369 |
| ALT-OPN-WTN | 2678 | 0,041111 | | | |
| ALT-OPN | 732 | 0,011237 | | | |
| WRKSP-CEL | 16003 | 0,921248 | **G2:Cellular Workspaces**<br>Tables, sofas or booths in an enclosed space, usually office spaces of executives that receive guest and therefore need more privacy or small teams. These can be alternative spots close to the main work area or further away with no visual connection. Included are also semi/enclosed with point with books/magazines and quire rooms. | 17371 | 0,098 |
| ALT-CLD-WTN | 1133 | 0,065224 | | | |
| ALT-CLD-AWY | 154 | 0,008865 | | | |
| ALT-CEL | 29 | 0,001669 | | | |
| ALT-LIB | 26 | 0,001497 | | | |
| ALT-QUT | 26 | 0,001497 | | | |
| MTG-BKB | 19578 | 0,913025 | **G3:Meeting Spaces**<br>These include bookable and non-bookable enclosed spaces, flexible alternative spaces and videoconferencing rooms. What unifies these rooms are the fact that they are shared enclosed and allow for temporary use. | 21443 | 0,122 |
| MTG-OTH-TRN | 968 | 0,045143 | | | |
| MTG-NBK | 426 | 0,019867 | | | |
| ALT-FLX | 384 | 0,017908 | | | |
| OTHFCL-AVC | 50 | 0,002332 | | | |
| MTG | 37 | 0,001726 | | | |
| OTHFCL-STO-CLD | 1697 | 0,204507 | **G4:Storage Facilites**<br>All storage types are considered: any height or enclosure. Included are also server and plant rooms. | 8298 | 0,047 |
| OTHFCL-STO-HGT | 993 | 0,119667 | | | |
| OTHFCL-CMM | 1716 | 0,206797 | | | |
| OTHFCL-STO | 3892 | 0,469029 | | | |
| OTHFCL-CAN-SIT | 3106 | 0,381713 | **G5:Kitchen/Tea Seating**<br>These include canteen areas and explicitly their seating area. Also open and closed tea spaces. | 8137 | 0,046 |
| OTHFCL-CAN | 2239 | 0,275163 | | | |
| OTHFCL-TEA-OPN | 1154 | 0,141821 | | | |
| OTHFCL-TEA-CLD | 905 | 0,11122 | | | |
| OTHFCL-TEA | 733 | 0,090082 | | | |
| OTHFCL-CAN-KTN | 1086 | 0,479682 | **G6:Kitchen/Tea Service**<br>These include catering kitchens, canteen serving area and externally run tea/coffee/vending stations. | 2264 | 0,013 |
| OTHFCL-CAN-SRV | 1025 | 0,452739 | | | |
| OTHFCL-TEA-CAT | 153 | 0,06758 | | | |
| OTHFCL-PRC-REP | 964 | 0,527064 | **G7:Print/Copy Facilities**<br>Any space used for printing, coping and in some cases, cutting mounting or sticking. | 1829 | 0,010 |
| OTHFCL-PRC-AWY | 185 | 0,101148 | | | |
| OTHFCL-PRC-WTH | 75 | 0,041006 | | | |
| OTHFCL-PRC | 605 | 0,330782 | | | |
| OTHFCL-REC-WTA | 651 | 0,203883 | **G8:Reception**<br>This includes the reception and the small circulation space around it and its furniture that is not primary. Pigeonholes and other functional spaces that cannot be used for working are also considered here. | 3193 | 0,018 |
| OTHFCL-REC | 1127 | 0,35296 | | | |
| OTHFCL-PST | 1124 | 0,35202 | | | |
| ALT-OPN-AWY | 291 | 0,091137 | | | |
| OTHFCL-GYM | 317 | 0,856757 | **G9:Extra_Facilites**<br>In the case studies examined there are a gym and a nursery. But they can be generalised to any miscellaneous, permanent facility, not directly linked to the activities in the office. | 370 | 0,003 |
| OTHFCL-NUR | 53 | | | | |
| CIRC-PRI | 35282 | 1 | **G10:Primary_Circulation** | 35282 | 0,200 |
| ? | 9544 | 1 | **G10:Secondary_Circulation** | 9544 | 0,054 |
| EXCLUDE | 4 | 0,001135 | **EXCLUDE:** These include spaces that don't relate to the offices or storage below eye sight (furniture). | 3524 | 0,020 |
| OTHFCL-STO-LOW | 3520 | 0,998865 | | | |

## Plan specific attributes

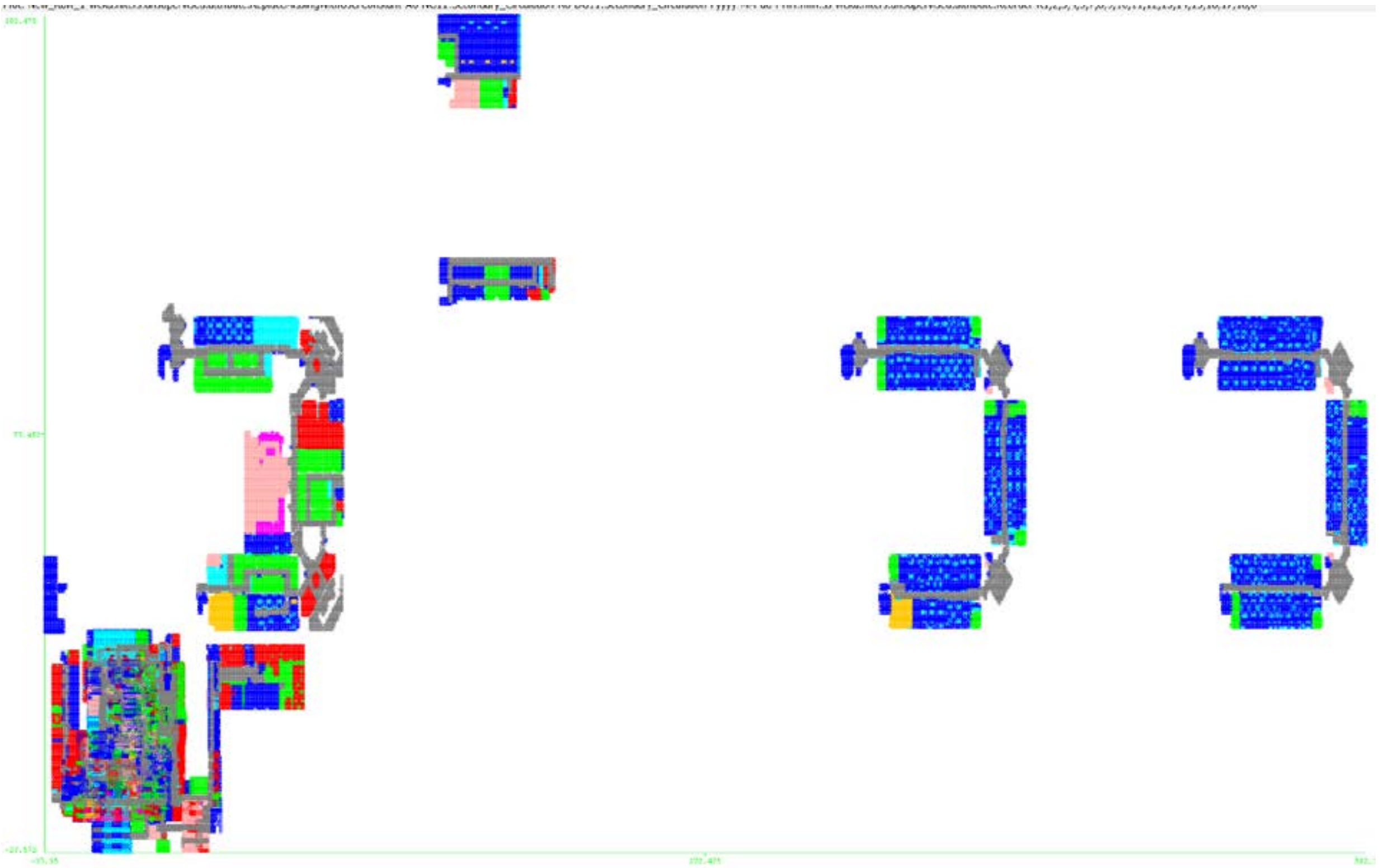

*Figure A3: Instances plotted by X/Y value to show that these are plan specific*

# Appendix B:

## OneR

A sample from the 2899 rules made from OneR based on visual mean depth. These achieved 127 140 correctly classified instances using 10-fold cross validation on the dataset of 176 395 instances. Although the rules are long because they have to cover a large data span with a high accuracy. There is an obvious trade-off between rule length and accuracy. Rule length can be controlled by increasing or decreasing the minimum amount of instances to be covered by each condition (in this case 6):

```
Visual Mean Depth:
    < 1.4490695 -> G10:Primary_Circulation
    < 1.4702685 -> G1:Open_Workspaces
    < 1.4741469999999999    -> G10:Primary_Circulation
    < 1.4806105 -> G1:Open_Workspaces
    < 1.4844879999999998    -> G10:Primary_Circulation
    < 1.5235265 -> G1:Open_Workspaces
    < 1.5405894999999998    -> G10:Primary_Circulation
    < 1.5498965 -> G1:Open_Workspaces
    < 1.557394  -> G10:Primary_Circulation
    < 1.5620479999999999    -> G1:Open_Workspaces
    < 1.5687695000000001    -> G10:Primary_Circulation
    < 1.5731644999999999    -> G1:Open_Workspaces
    < 1.5793694999999999    -> G10:Primary_Circulation
    < 1.5824715 -> G1:Open_Workspaces
    < 1.5904859999999998    -> G10:Primary_Circulation
    < 1.598242  -> G1:Open_Workspaces
    < 1.6072905 -> G10:Primary_Circulation
    < 1.6184075 -> G1:Open_Workspaces
    < 1.6233195 -> G10:Primary_Circulation
    < 1.6295245 -> G1:Open_Workspaces
    < 1.6331435 -> G10:Primary_Circulation
    < 1.6377975 -> G1:Open_Workspaces
    < 1.6419335 -> G2:Cellular_Workspaces
    < 1.6481385 -> G10:Primary_Circulation
    < 1.6522755 -> G1:Open_Workspaces
    < 1.6641675 -> G10:Primary_Circulation
    < 1.6786455 -> G1:Open_Workspaces
    < 1.6848505 -> G10:Primary_Circulation
    < 1.6879525 -> G8:Reception
    < 1.691313  -> G1:Open_Workspaces
    < 1.6962255000000002    -> G8:Reception
    < 1.7070835 -> G1:Open_Workspaces
    < 1.7101864999999998    -> G10:Primary_Circulation
    < 1.7225955000000002    -> G1:Open_Workspaces
    < 1.7259570000000002    -> G8:Reception
    < 1.7331955 -> G2:Cellular_Workspaces
    < 1.759566  -> G8:Reception
    < 1.7662875 -> G1:Open_Workspaces
    < 1.7753364999999999    -> G3:Meeting_Spaces
    < 1.7929165 -> G8:Reception
    < 1.9092555 -> G1:Open_Workspaces
    < 2.105481  -> G3:Meeting_Spaces
    < 2.146846  -> G5:Kitchen/Tea_Seating

                                  ……………………………………………………………………

    < 11.4749975    -> G4:Storage_Facilites
    < 11.5442695    -> G3:Meeting_Spaces
    < 11.625922500000001    -> G6:Kitchen/Tea_Service
    < 11.805269 -> G1:Open_Workspaces
    >= 11.805269    -> G11:Secondary_Circulation
(127140/176395 instances correct)
```

*Figure B1: Sample of OneR rules*

# Appendix C:

## J48

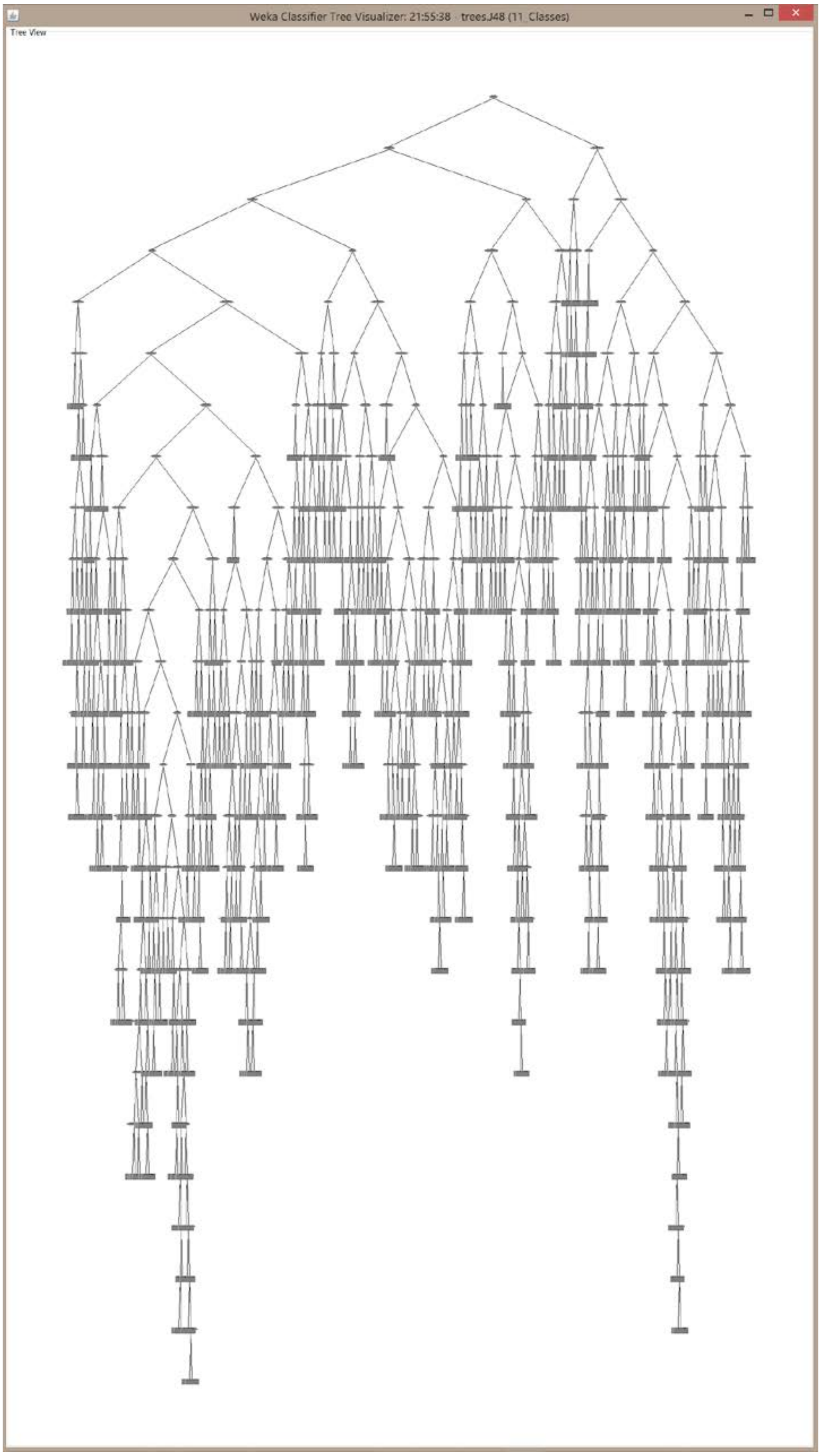

*Figure C1: Visualisation of the J48 Decision tree with EBP*

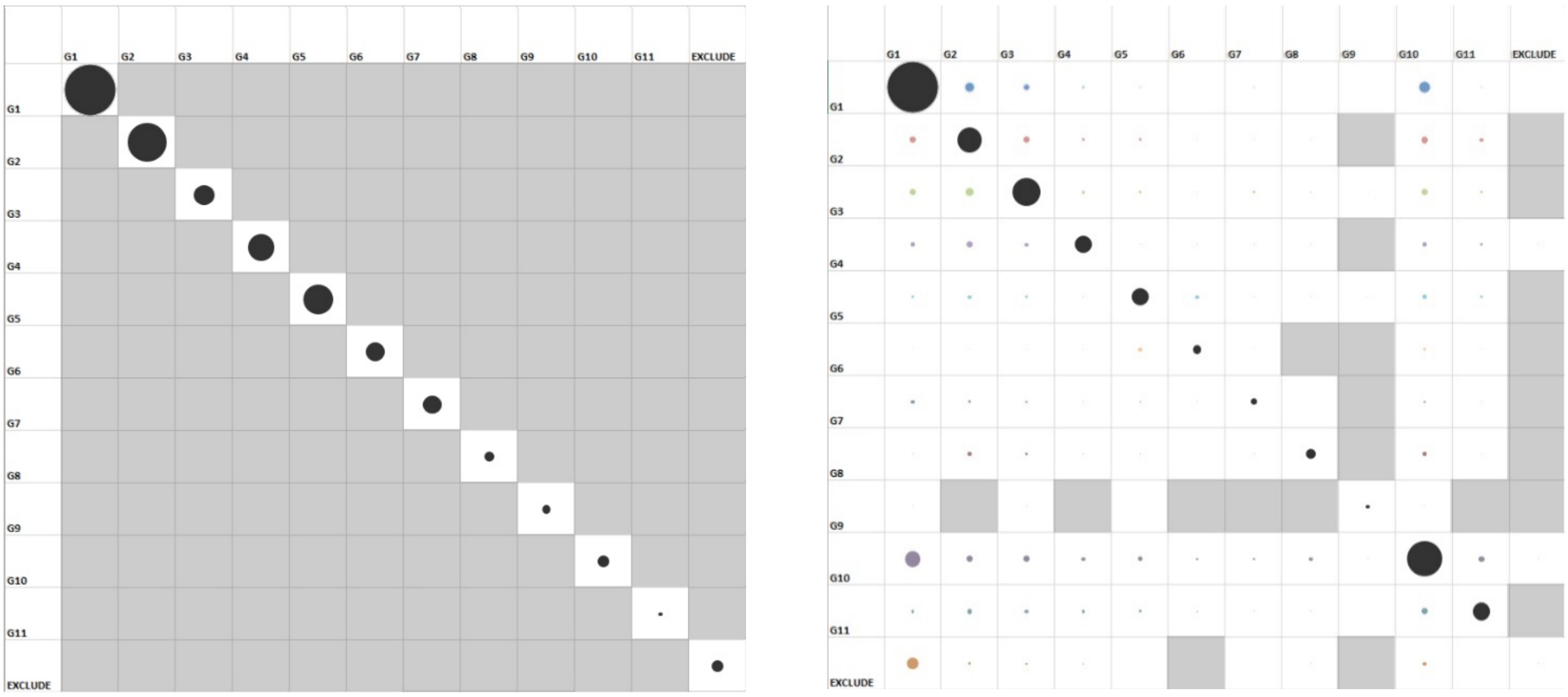

*Visualisation of an ideal confusion matrix heat map to that of J48 Decision tree with EBP*

## Confusion Matrixes

| | G1 | G2 | G3 | G4 | G5 | G6 | G7 | G8 | G9 | G10 | G11 | EXCLUDE |
|---|---|---|---|---|---|---|---|---|---|---|---|---|
| **G1** | 58266 | 2230 | 904 | 305 | 76 | 11 | 78 | 15 | 10 | 3106 | 132 | 7 |
| **G2** | 1058 | 13694 | 933 | 106 | 102 | 9 | 48 | 35 | | 1080 | 306 | |
| **G3** | 938 | 1464 | 17460 | 276 | 117 | 11 | 95 | 59 | 4 | 907 | 112 | |
| **G4** | 479 | 927 | 299 | 5960 | 15 | 21 | 20 | 4 | | 400 | 140 | 33 |
| **G5** | 100 | 290 | 186 | 4 | 6709 | 318 | 13 | 33 | 2 | 386 | 96 | |
| **G6** | 3 | 6 | 25 | 6 | 328 | 1753 | 11 | | | 100 | 32 | |
| **G7** | 312 | 204 | 95 | 32 | 62 | 11 | 974 | 1 | | 97 | 41 | |
| **G8** | 31 | 433 | 114 | 15 | 24 | 1 | 13 | 1987 | | 540 | 35 | |
| **G9** | 9 | | 6 | | 1 | | | | 331 | 23 | | |
| **G10** | 5383 | 941 | 882 | 303 | 477 | 92 | 72 | 329 | 9 | 26104 | 682 | 8 |
| **G11** | 267 | 630 | 329 | 210 | 149 | 58 | 31 | 26 | 4 | 929 | 6911 | |
| **EXCLUDE** | 2827 | 160 | 97 | 38 | 3 | | 2 | 20 | | 347 | 6 | 24 |

*J48 Complete set with EBP*

| | G1 | G2 | G3 | G4 | G5 | G6 | G7 | G8 | G9 | G10 | G11 | EXCLUDE |
|---|---|---|---|---|---|---|---|---|---|---|---|---|
| **G1** | 58323 | 2373 | 854 | 231 | 134 | 7 | 52 | 33 | 6 | 3011 | 116 | |
| **G2** | 1250 | 13475 | 866 | 111 | 91 | 9 | 74 | 28 | 2 | 1133 | 332 | |
| **G3** | 1097 | 1568 | 16940 | 239 | 211 | 15 | 74 | 76 | 23 | 1033 | 167 | |
| **G4** | 560 | 914 | 348 | 5711 | 27 | 18 | 38 | 9 | 2 | 374 | 291 | 6 |
| **G5** | 130 | 328 | 283 | 20 | 6293 | 279 | 11 | 26 | | 630 | 137 | |
| **G6** | 10 | 13 | 14 | 15 | 371 | 1701 | 5 | | | 76 | 59 | |
| **G7** | 356 | 226 | 101 | 33 | 88 | | 879 | | | 103 | 43 | |
| **G8** | 29 | 430 | 121 | 5 | 47 | | 10 | 1844 | | 672 | 35 | |
| **G9** | 21 | | 22 | | 1 | | | | 307 | 19 | | |
| **G10** | 6279 | 1079 | 1045 | 335 | 431 | 120 | 41 | 309 | 9 | 24898 | 736 | |
| **G11** | 353 | 665 | 302 | 183 | 156 | 55 | 35 | 23 | 5 | 960 | 6807 | |
| **EXCLUDE** | 2869 | 169 | 93 | 42 | 12 | 1 | 2 | 11 | | 307 | 5 | 13 |

*J48 Complete set with REP*

| | G1 | G2 | G3 | G4 | G5 | G6 | G7 | G8 | G9 | G10 | G11 | EXCLUDE |
|---|---|---|---|---|---|---|---|---|---|---|---|---|
| **G1** | 58274 | 2213 | 947 | 259 | 61 | 11 | 70 | 21 | 15 | 3151 | 111 | 7 |
| **G2** | 1112 | 13598 | 979 | 84 | 122 | 15 | 78 | 32 | | 1055 | 296 | |
| **G3** | 961 | 1407 | 17544 | 248 | 165 | 7 | 78 | 40 | 14 | 867 | 112 | |
| **G4** | 521 | 885 | 365 | 5813 | 30 | 6 | 31 | 16 | | 427 | 195 | 9 |
| **G5** | 133 | 259 | 194 | 9 | 6640 | 324 | 19 | 30 | 1 | 427 | 101 | |
| **G6** | 3 | | 6 | | 328 | 1784 | 10 | | | 72 | 61 | |
| **G7** | 325 | 218 | 93 | 49 | 74 | | 941 | 1 | | 93 | 35 | |
| **G8** | 31 | 412 | 129 | 2 | 19 | 1 | | 2038 | | 525 | 36 | |
| **G9** | 1 | | 7 | | 1 | | | | 348 | 13 | | |
| **G10** | 5484 | 1004 | 924 | 265 | 442 | 70 | 64 | 340 | 9 | 26055 | 617 | 8 |
| **G11** | 304 | 666 | 312 | 170 | 133 | 48 | 26 | 28 | 3 | 969 | 6885 | |
| **EXCLUDE** | 2822 | 157 | 95 | 34 | 3 | | 2 | 18 | | 363 | 5 | 25 |

*J48 Subset with EBP*

| | G1 | G2 | G3 | G4 | G5 | G6 | G7 | G8 | G9 | G10 | G11 | EXCLUDE |
|---|---|---|---|---|---|---|---|---|---|---|---|---|
| **G1** | 58372 | 2518 | 926 | 206 | 171 | 8 | 35 | 50 | 8 | 2734 | 110 | 2 |
| **G2** | 1443 | 12946 | 993 | 142 | 196 | 7 | 40 | 16 | 2 | 1232 | 354 | |
| **G3** | 1310 | 1519 | 16833 | 235 | 269 | 7 | 75 | 88 | 1 | 961 | 145 | |
| **G4** | 618 | 901 | 417 | 5484 | 15 | 20 | 59 | 13 | | 405 | 357 | 9 |
| **G5** | 163 | 355 | 256 | 27 | 6452 | 215 | 30 | 27 | 2 | 485 | 125 | |
| **G6** | 11 | 11 | 8 | 2 | 508 | 1577 | | | | 118 | 26 | 3 |
| **G7** | 402 | 240 | 138 | 20 | 58 | | 810 | | | 118 | 43 | |
| **G8** | 67 | 452 | 132 | 8 | 12 | | 4 | 1876 | | 590 | 52 | |
| **G9** | 17 | | | 3 | | | | | 299 | 6 | 45 | |
| **G10** | 7013 | 1140 | 1127 | 370 | 529 | 89 | 62 | 311 | 11 | 23781 | 839 | 10 |
| **G11** | 388 | 747 | 377 | 169 | 213 | 49 | 20 | 19 | 4 | 1065 | 6484 | 9 |
| **EXCLUDE** | 2935 | 158 | 94 | 32 | 8 | | | 15 | | 238 | 7 | 37 |

*Subset with REP*

| | G1 | G2 | G3 | G4 | G5 | G6 | G7 | G8 | G9 | G10 | G11 | EXCLUDE |
|---|---|---|---|---|---|---|---|---|---|---|---|---|
| **G1** | 58529 | 2351 | 780 | 231 | 101 | 14 | 37 | 46 | 9 | 2954 | 86 | 2 |
| **G2** | 1402 | 13362 | 888 | 116 | 78 | | 63 | 53 | 1 | 1109 | 299 | |
| **G3** | 1225 | 1503 | 17147 | 254 | 162 | 4 | 81 | 62 | 10 | 886 | 109 | |
| **G4** | 593 | 857 | 290 | 5818 | 25 | 9 | 16 | 26 | | 405 | 241 | 18 |
| **G5** | 112 | 315 | 158 | 4 | 6703 | 283 | 8 | 23 | 3 | 437 | 91 | |
| **G6** | 2 | 2 | 19 | 12 | 344 | 1761 | | | | 69 | 42 | 13 |
| **G7** | 404 | 187 | 132 | 46 | 65 | | 871 | | | 81 | 43 | |
| **G8** | 64 | 440 | 90 | 5 | 18 | | 2 | 2009 | | 535 | 30 | |
| **G9** | 6 | | | 1 | | | | | 354 | 7 | 2 | |
| **G10** | 6195 | 1108 | 889 | 276 | 403 | 95 | 39 | 301 | 23 | 25214 | 711 | 28 |
| **G11** | 407 | 674 | 282 | 156 | 169 | 22 | 29 | 36 | 14 | 1020 | 6722 | 13 |
| **EXCLUDE** | 2923 | 151 | 88 | 6 | 3 | | | 16 | | 279 | 4 | 54 |

*PCA with EBP*

| | G1 | G2 | G3 | G4 | G5 | G6 | G7 | G8 | G9 | G10 | G11 | EXCLUDE |
|---|---|---|---|---|---|---|---|---|---|---|---|---|
| **G1** | 58211 | 2503 | 821 | 173 | 126 | 11 | 45 | 33 | 11 | 3094 | 102 | 10 |
| **G2** | 1172 | 13493 | 910 | 81 | 86 | 8 | 55 | 31 | | 1173 | 358 | 4 |
| **G3** | 1153 | 1565 | 16849 | 211 | 205 | 8 | 84 | 51 | 24 | 1102 | 191 | |
| **G4** | 561 | 991 | 361 | 5607 | 27 | 29 | 31 | 17 | 4 | 347 | 318 | 5 |
| **G5** | 146 | 298 | 303 | 13 | 6299 | 281 | 20 | 30 | | 601 | 146 | |
| **G6** | 5 | 3 | 18 | 22 | 375 | 1717 | 2 | 1 | | 50 | 71 | |
| **G7** | 378 | 212 | 69 | 37 | 90 | 10 | 883 | | | 101 | 49 | |
| **G8** | 31 | 428 | 106 | 7 | 32 | 1 | 3 | 1881 | | 669 | 35 | |
| **G9** | 14 | | 17 | | 1 | | | | 318 | 20 | | |
| **G10** | 6203 | 1120 | 1049 | 321 | 450 | 120 | 24 | 332 | 17 | 24926 | 718 | 2 |
| **G11** | 359 | 649 | 297 | 178 | 147 | 28 | 23 | 24 | 5 | 964 | 6868 | 2 |
| **EXCLUDE** | 2862 | 163 | 100 | 50 | 11 | 1 | 1 | 10 | | 313 | 5 | 8 |

*PCA with REP*

## Appendix D:

*K-Means Clustering Overview*

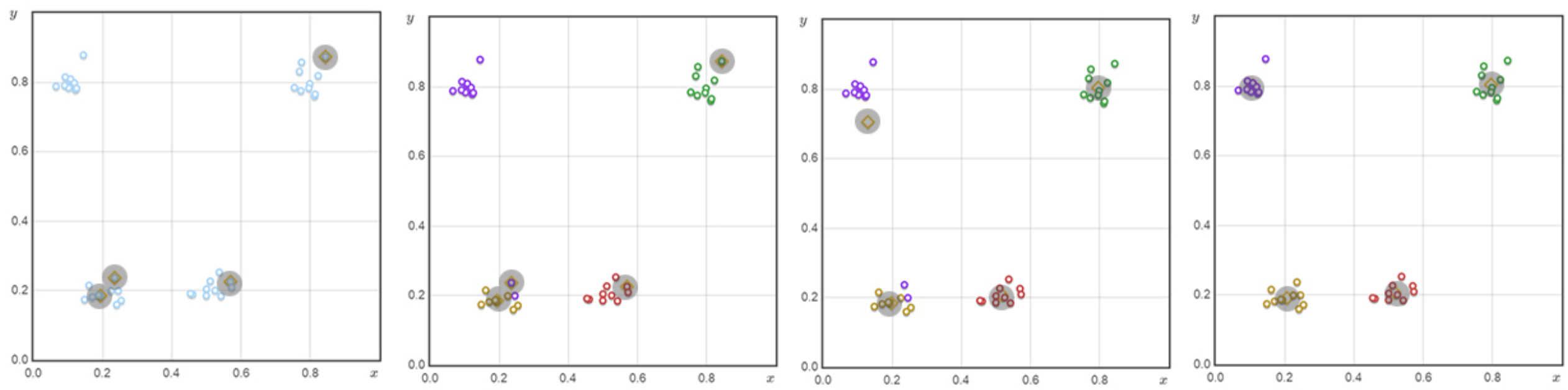

*Figure D1 (based on Hugo, 2014): K-Means clustering with 4 clusters. Note how the centroid of the purple cluster moves in the third iteration to be closer to the points that belong to it*

Clustering is a type of unsupervised learning that orders instances in a dataset without regards to class labels, but rather only by “natural groupings” - intrinsic patterns in the dataset (Jain, 2009 p. 652). One of the most popular and prevailing algorithm for clustering is K-means, which has been introduced independently, multiple times in the 1950s (Wu et al., 2007; Jain, 2009 p. 653). K-means clustering is done by pre-specifying ‘k’ number of wanted clusters and assigning ‘k’ random instances of the dataset to be centroids for these clusters (depicted as grey dots in Figure 12). Afterwards, the distance of every instance in the dataset to all the cluster centroids is computed, either by using Euclidean or Manhattan distance. The instances are said to belong to the k-cluster centroid nearest to it. After all the instances that belong to a cluster have been considered, the centroid is recalculated so that is represents the average of the cluster. The instances-to-cluster assignment procedure and centroid recalibration is repeated iteratively until the learner converges (see flowchart in figure D2).

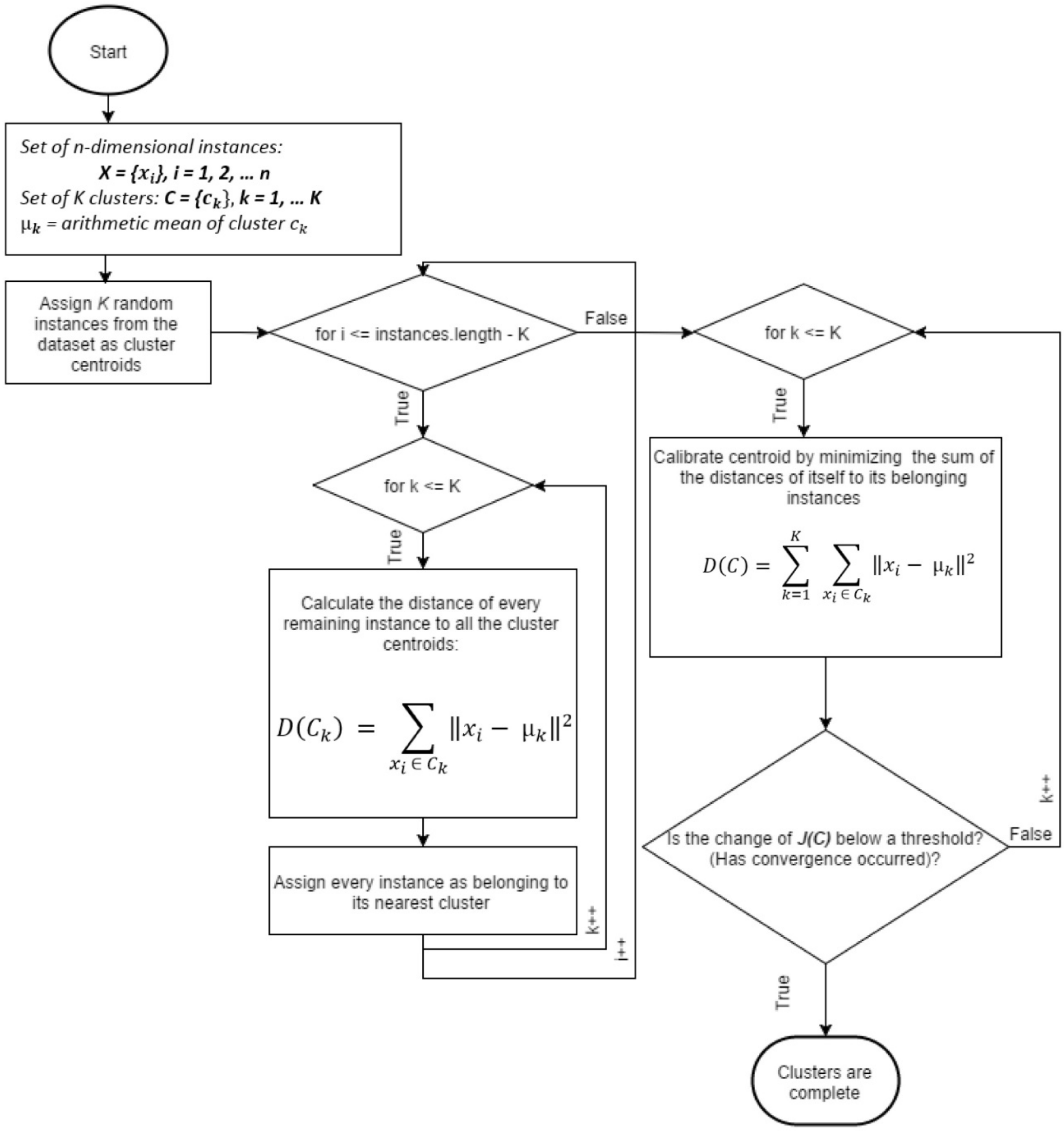

*Figure D2: K-Means flowchart (Formulas based on Jain, 2009 p. 654)*

## *Confusion Matrixes*

| | G1 | G2 | G3 | G4 | G5 | G6 | G7 | G8 | G9 | G10 | G11 | EXCLUDE |
|---|---|---|---|---|---|---|---|---|---|---|---|---|
| **G1** | 804 | 2996 | 7776 | 7333 | 6060 | 8513 | 664 | 1938 | 4398 | 4137 | 9472 | 11049 |
| **G2** | 143 | 6548 | | 2600 | | 7288 | 98 | 203 | 307 | | 184 | |
| **G3** | 1610 | 1516 | 65 | 4266 | 260 | 6249 | 453 | 4934 | 1259 | 250 | 545 | 36 |
| **G4** | 2 | 1471 | | 749 | 246 | 2476 | 17 | 1234 | 1965 | | 138 | |
| **G5** | 234 | 267 | 330 | 3173 | 12 | 1281 | 107 | 1518 | 46 | | 1169 | |
| **G6** | | 26 | | 1156 | | 1 | | 501 | | | 580 | |
| **G7** | 6 | 244 | 5 | 312 | 40 | 175 | 12 | 489 | 129 | 21 | 48 | 348 |
| **G8** | 917 | 183 | | 461 | | 934 | 102 | 489 | 107 | | | |
| **G9** | | | | | | 317 | | 53 | | | | |
| **G10** | 7149 | 2284 | 1605 | 5280 | 692 | 9495 | 386 | 1750 | 1751 | 546 | 2962 | 1382 |
| **G11** | 1153 | 1826 | 19 | 2560 | | 2166 | 8 | 1511 | 215 | 12 | 70 | 4 |
| **EXCLUDE** | 89 | 134 | 1124 | 37 | 43 | 104 | | 273 | 97 | 429 | | 1194 |

*K-Means on complete data and attribute set*

| | G1 | G2 | G3 | G4 | G5 | G6 | G7 | G8 | G9 | G10 | G11 | EXCLUDE |
|---|---|---|---|---|---|---|---|---|---|---|---|---|
| **G1** | 803 | 2986 | 7737 | 4084 | 6052 | 7092 | 5476 | 1936 | 4315 | 4143 | 9431 | 11085 |
| **G2** | 143 | 6478 | | 1604 | | 6605 | 2015 | 203 | 144 | | 179 | |
| **G3** | 1610 | 1290 | 65 | 3343 | 247 | 4934 | 3104 | 4934 | 1086 | 250 | 544 | 36 |
| **G4** | 2 | 1468 | | 1361 | 237 | 1835 | 440 | 1234 | 1586 | | 135 | |
| **G5** | 234 | 267 | 385 | 2060 | 12 | 1159 | 1386 | 1463 | 46 | | 1125 | |
| **G6** | | 1 | 2 | 92 | | 26 | 1169 | 499 | | | 475 | |
| **G7** | 6 | 244 | 5 | 70 | 40 | 170 | 268 | 489 | 128 | 21 | 40 | 348 |
| **G8** | 917 | 183 | | 571 | | 507 | 419 | 489 | 107 | | | |
| **G9** | | | | 40 | | 277 | | 53 | | | | |
| **G10** | 7150 | 2200 | 1589 | 4196 | 680 | 7214 | 4063 | 1747 | 1696 | 546 | 2801 | 1400 |
| **G11** | 1153 | 1668 | 19 | 2107 | | 1385 | 1453 | 1511 | 167 | 12 | 65 | 4 |
| **EXCLUDE** | 89 | 134 | 1122 | 32 | 43 | 90 | 21 | 273 | 95 | 429 | | 1196 |

*K-Means on complete data and attribute subset*

| | G1 | G2 | G3 | G4 | G5 | G6 | G7 | G8 | G9 | G10 | G11 | EXCLUDE |
|---|---|---|---|---|---|---|---|---|---|---|---|---|
| **G1** | 2241 | 3610 | 11692 | 5223 | 632 | 4384 | 10881 | 1735 | 5166 | 10036 | 653 | 8887 |
| **G2** | 140 | 5133 | | 1976 | 28 | 7956 | 1457 | 206 | 386 | | 89 | |
| **G3** | 1224 | 3245 | 71 | 3714 | 29 | 1695 | 2667 | 5320 | 2644 | 280 | 117 | 437 |
| **G4** | 297 | 2302 | | 270 | 19 | 1515 | 433 | 939 | 2195 | | 17 | 311 |
| **G5** | 313 | 1339 | | 258 | 1340 | 358 | 1019 | 1769 | 1712 | | 5 | 24 |
| **G6** | | 1 | | 686 | 1050 | 26 | | 501 | | | | |
| **G7** | 482 | 110 | 93 | 58 | 6 | 272 | 317 | 13 | 132 | 281 | 11 | 54 |
| **G8** | 546 | 232 | | 821 | | 195 | 194 | 860 | 245 | | 100 | |
| **G9** | | 317 | | | | | | 53 | | | | |
| **G10** | 6970 | 5769 | 2016 | 4935 | 708 | 2924 | 4241 | 2281 | 2422 | 1165 | 382 | 1469 |
| **G11** | 1844 | 1289 | 19 | 1676 | 29 | 1930 | 1191 | 820 | 720 | 16 | 8 | 2 |
| **EXCLUDE** | 496 | 80 | 1405 | 40 | | 142 | 8 | 162 | 102 | 1046 | | 43 |

*K-Means on complete data and PCA attributes*

| | G1 | G2 | G3 | G4 | G5 | G6 | G7 | G8 | G9 | G10 | G11 | EXCLUDE |
|---|---|---|---|---|---|---|---|---|---|---|---|---|
| **G1** | 723 | 3288 | 5131 | 1 | 2262 | 1127 | 6086 | 3 | 1545 | | 4723 | 815 |
| **G2** | 141 | | | 2 | | 203 | | | | | | |
| **G3** | 1414 | 11 | 53 | 7 | 200 | 2111 | 64 | 24 | | 2374 | 23 | 614 |
| **G4** | 1 | | | | | 786 | | 1 | | 177 | | 271 |
| **G5** | 228 | | | 6 | | 1763 | | | | 3 | | 82 |
| **G6** | | | | | | 467 | | | | 34 | | |
| **G7** | 6 | 113 | 13 | | | 10 | 4 | | 1 | 237 | 243 | 242 |
| **G8** | 413 | | | 504 | | 489 | | | | | | |
| **G9** | | | | | | 52 | | | | 1 | | |
| **G10** | 2712 | 606 | 410 | 2666 | 347 | 676 | 1161 | 1678 | 609 | 258 | 675 | 634 |
| **G11** | 419 | 4 | | 32 | 10 | 204 | 17 | 447 | 2 | 621 | 2 | 941 |
| **EXCLUDE** | 81 | 438 | 468 | | 249 | 95 | 778 | | 350 | 10 | 481 | 159 |

*K-Means on single largest layout with complete dataset*

| | G1 | G2 | G3 | G4 | G5 | G6 | G7 | G8 | G9 | G10 | G11 | EXCLUDE |
|---|---|---|---|---|---|---|---|---|---|---|---|---|
| **G1** | 128 | 4260 | 5445 | | 2028 | 1097 | 6153 | 112 | 2231 | 8 | 3613 | 629 |
| **G2** | 138 | | | 2 | | 202 | | | | | | 4 |
| **G3** | 19 | 32 | 71 | | 184 | 1610 | 64 | 513 | | 2295 | | 2107 |
| **G4** | | | | | | 775 | | 248 | | 152 | | 61 |
| **G5** | 182 | | | 6 | | 1568 | | 1 | | 81 | | 244 |
| **G6** | | | | | | 501 | | | | | | |
| **G7** | 3 | 309 | 61 | | | 10 | 4 | 7 | | 472 | | 3 |
| **G8** | 24 | | | 504 | | 489 | | | | | | 389 |
| **G9** | | | | | | 52 | | | | | | 1 |
| **G10** | 2325 | 650 | 530 | 2157 | 333 | 383 | 1158 | 1787 | 618 | 357 | 524 | 1610 |
| **G11** | 328 | 2 | | 25 | 12 | 85 | 17 | 1276 | 2 | 720 | 2 | 230 |
| **EXCLUDE** | 33 | 502 | 507 | | 231 | 95 | 782 | 35 | 444 | 36 | 398 | 46 |

*K-Means on single largest layout with attributes subset*

*Self-Organising Maps Overview*

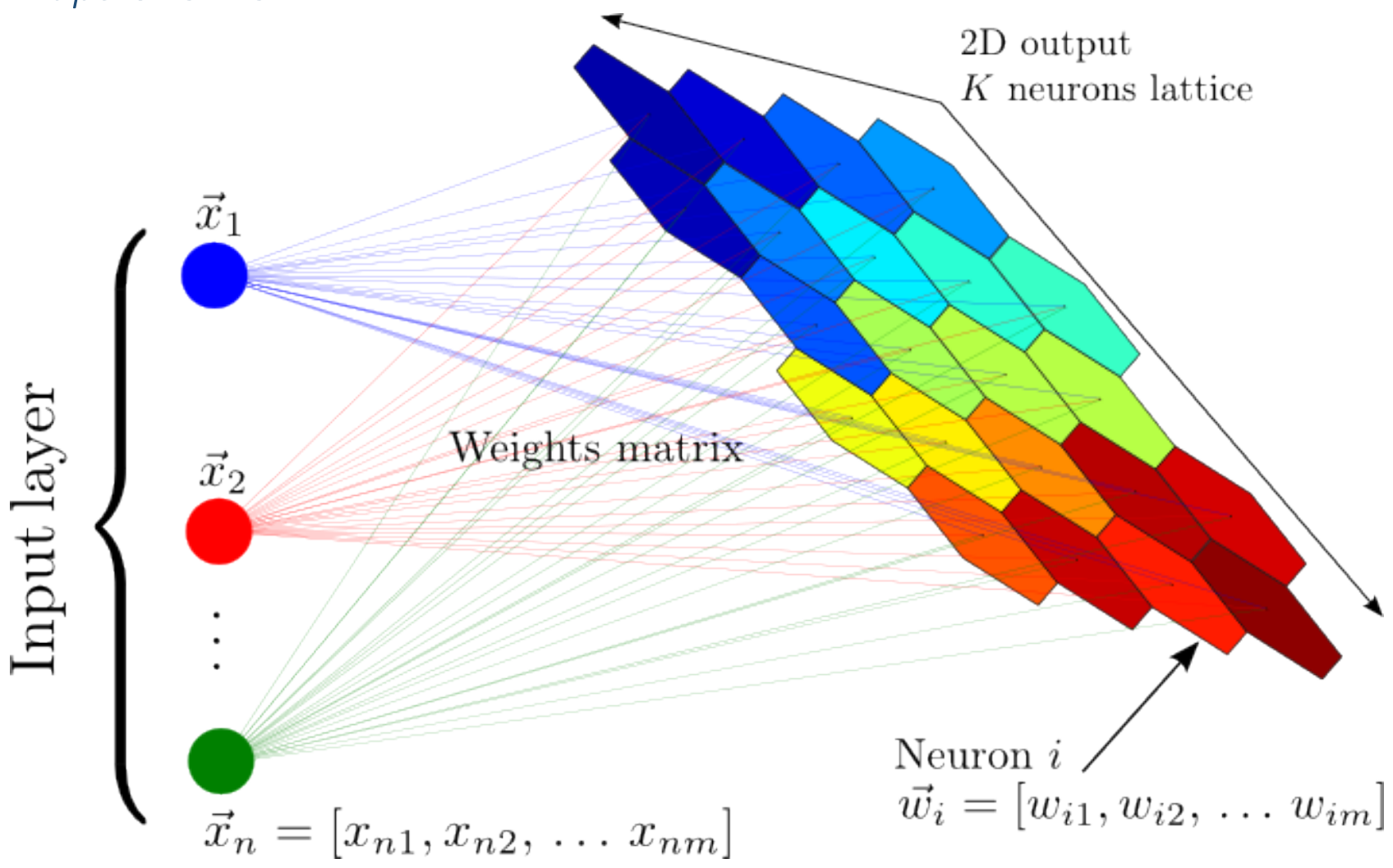

*Figure D3 (based on Carrasco Kind and Brunner, 2013 p. 3411): Visualization of a Kohonen map: Every node in the input layer is connected with all neurons in the competitive layer by a weight vector. The nodes in the competitive layer are wired to input node, but also to their own neighbours.*

Self-organising maps (SOM) or Kohonen maps are a competitive learning algorithm developed in the 1980s based on artificial neural networks, which are intended to model learning by association as done in the human brain. The idea of this algorithms that neurons that ‘fire together wire together’ (Shatz, 1992 p.21). SOMs consist of an input layer connected to a competitive layer, initially with random weight vectors which get adjusted over run time (Salatas, 2011; Kohonen, 1982). When an n-dimensional instance is introduced to the n-number of input nodes, the neurons in the competitive layer are triggered to look for the weight vector most similar to the input vector based on Euclidean distance. Once the best matching vector is found it is chosen as a winner and the weights to its neighbouring neurons are strengthened to react stronger to that particular input (see Figure 14). This strengthening of connected weight vectors allows the SOM to map an ‘n’ dimensional input to the 2d-lattice of the competitive layer (refer to image 14).

$$weight(t+1) = weight(t) + (Learning\ Rate)(Neighbourhoood\ impact\ rate)\left(\overrightarrow{Input} - \overrightarrow{weight(t)}\right)$$

*Figure D4: Simplified formula description for the SOM weights update function. Note that as time increases, so does neighbourhood impact and learning rate decrease*

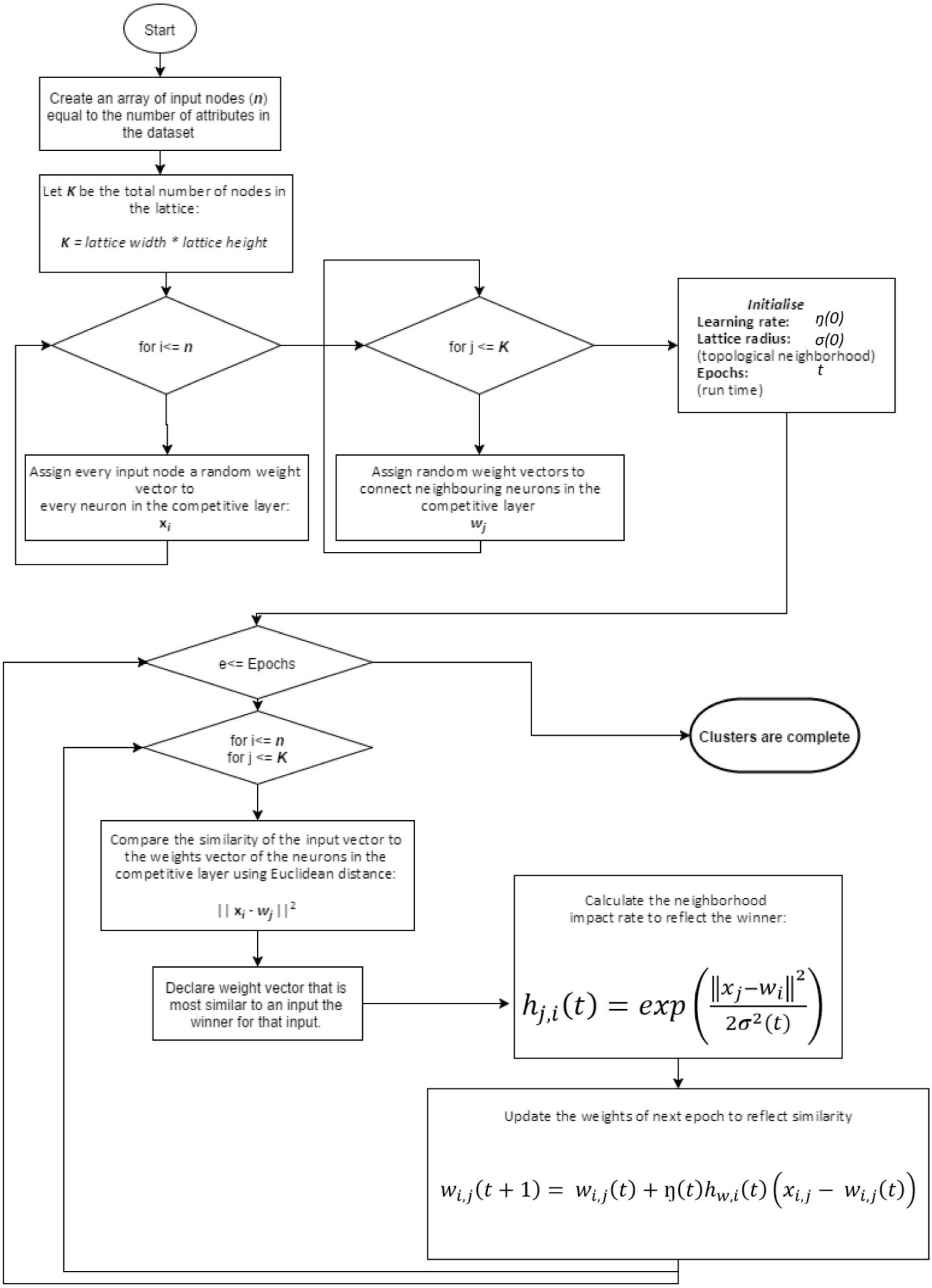

*Figure D5: SOM flowchart (Formulas based on Kind and Brunner, 2014 and Ai-junkie.com, 2010)*

## *PCA of the Single Layout*

| V1 | V2 | V3 | |
|---|---|---|---|
| 0.4182 | 0.4023 | 0.1086 | Point First Moment |
| -0.4326 | 0.3681 | 0.0399 | Visual Mean Depth |
| 0.4277 | -0.3834 | -0.0278 | Visual Integration [Tekl] |
| -0.1289 | 0.4196 | -0.7699 | Visual Entropy |
| -0.338 | 0.2854 | 0.6061 | Visual Relativised Entropy |
| 0.3841 | 0.4362 | 0.0819 | Point Second Moment |
| 0.4203 | 0.3284 | 0.1379 | Connectivity |

## *Confusion Matrixes*

| | **G1** | **G2** | **G3** | **G4** | **G5** | **G6** | **G7** | **G8** | **G9** | **G10** | **G11** | **EXCLUDE** |
|---|---|---|---|---|---|---|---|---|---|---|---|---|
| **G1** | 2239 | 664 | 8863 | 5271 | 7814 | 332 | 4062 | 20433 | 5519 | 2547 | 3752 | 3644 |
| **G2** | | 98 | 5715 | | 185 | 145 | 1848 | 346 | | 999 | 4483 | 3552 |
| **G3** | 159 | 283 | 7455 | 252 | 411 | 170 | 2683 | 6643 | 135 | 1100 | 1150 | 1002 |
| **G4** | 51 | 17 | 3436 | | 131 | 135 | 267 | 1236 | 85 | 240 | 1768 | 932 |
| **G5** | | 94 | 2937 | | 1148 | 279 | 1114 | 2082 | | 26 | 102 | 355 |
| **G6** | | | 118 | | 543 | | 1101 | 501 | | | 1 | |
| **G7** | 8 | 12 | 222 | 40 | 49 | 37 | 213 | 829 | 30 | 15 | 275 | 99 |
| **G8** | | 102 | 999 | | | 59 | 311 | 1406 | | 36 | 155 | 125 |
| **G9** | | | 317 | | | | | 53 | | | | |
| **G10** | 609 | 384 | 8241 | 650 | 2471 | 294 | 3306 | 11782 | 595 | 1419 | 2214 | 3317 |
| **G11** | | 8 | 3972 | 12 | 74 | 207 | 1365 | 2687 | | 49 | 915 | 255 |
| **EXCLUDE** | | | 167 | 543 | | | 22 | 2566 | 43 | 28 | 90 | 65 |

*SOM on complete dataset and attributes*

| | **G1** | **G2** | **G3** | **G4** | **G5** | **G6** | **G7** | **G8** | **G9** | **G10** | **G11** | **EXCLUDE** |
|---|---|---|---|---|---|---|---|---|---|---|---|---|
| **G1** | 2246 | 664 | 5269 | 20435 | 7314 | 4786 | 1128 | 4927 | 5923 | 7731 | 365 | 4352 |
| **G2** | | 98 | | 346 | 183 | 1772 | | 1948 | | 6501 | 282 | 6241 |
| **G3** | 159 | 462 | 252 | 6643 | 393 | 2641 | | 1220 | 191 | 7142 | 215 | 2125 |
| **G4** | 54 | 17 | | 1236 | 89 | 245 | 11 | 1008 | 124 | 2881 | 206 | 2427 |
| **G5** | | 111 | | 2082 | 1141 | 1110 | | 427 | 10 | 2841 | 291 | 124 |
| **G6** | | | | 501 | 522 | 1122 | | | | 93 | | 26 |
| **G7** | 8 | 12 | 40 | 829 | 40 | 223 | 4 | 79 | 40 | 186 | 43 | 325 |
| **G8** | | 102 | | 1406 | | 311 | | 44 | | 912 | 59 | 359 |
| **G9** | | | | 53 | | | | 1 | | 316 | | |
| **G10** | 612 | 418 | 650 | 11782 | 2301 | 3492 | 208 | 4138 | 622 | 8065 | 288 | 2706 |
| **G11** | | 14 | 12 | 2687 | 69 | 1378 | 2 | 102 | | 3677 | 328 | 1275 |
| **EXCLUDE** | | | 543 | 2566 | | 21 | | 59 | 43 | 100 | 1 | 191 |

*SOM on complete dataset and attribute subset*

| | **G1** | **G2** | **G3** | **G4** | **G5** | **G6** | **G7** | **G8** | **G9** | **G10** | **G11** | **EXCLUDE** |
|---|---|---|---|---|---|---|---|---|---|---|---|---|
| **G1** | 13630 | 1202 | 5551 | 171 | 1770 | 2220 | 493 | 1008 | 14399 | 11534 | 9412 | 3750 |
| **G2** | 12239 | 246 | 936 | 9 | 66 | | 89 | | 3445 | | | 341 |
| **G3** | 5675 | 1271 | 891 | 394 | 243 | 173 | 68 | 47 | 5833 | 65 | 239 | 6544 |
| **G4** | 5402 | 72 | 220 | 11 | 24 | 50 | 6 | | 1277 | | | 1236 |
| **G5** | 2058 | 1176 | 4 | 105 | 1337 | | 2 | | 1373 | | | 2082 |
| **G6** | 27 | | | | 1490 | | | | 246 | | | 501 |
| **G7** | 506 | 67 | 41 | 6 | 16 | | 6 | | 318 | 45 | 328 | 496 |
| **G8** | 415 | 236 | 33 | 3 | 46 | | 99 | | 955 | | | 1406 |
| **G9** | 201 | | | | | | | | 116 | | | 53 |
| **G10** | 9021 | 1164 | 1486 | 24 | 1762 | 525 | 379 | 194 | 8489 | 1985 | 1045 | 9208 |
| **G11** | 3116 | 641 | 26 | 2 | 311 | | 8 | 2 | 2741 | 19 | 14 | 2664 |
| **EXCLUDE** | 330 | | 35 | | 7 | | | 99 | 43 | 1408 | 998 | 604 |

*SOM on complete dataset and PCA attributes*

| | G1 | G2 | G3 | G4 | G5 | G6 | G7 | G8 | G9 | G10 | G11 | EXCLUDE |
|---|---|---|---|---|---|---|---|---|---|---|---|---|
| **G1** | 5346 | 2179 | | 1581 | 1240 | 1180 | 128 | 295 | 483 | 2126 | 10095 | 1051 |
| **G2** | | | 2 | 206 | | | 138 | | | | | |
| **G3** | 1 | 4318 | | 2182 | | 64 | 15 | 29 | | | 215 | 71 |
| **G4** | | 1234 | | 1 | | | | 1 | | | | |
| **G5** | | 125 | 6 | 1769 | | | 126 | 56 | | | | |
| **G6** | | | | 501 | | | | | | | | |
| **G7** | 29 | 479 | | 13 | | 2 | 3 | | | | 343 | |
| **G8** | | 1 | 267 | 859 | | | 261 | 18 | | | | |
| **G9** | | 48 | | 5 | | | | | | | | |
| **G10** | 882 | 1803 | 1857 | 1677 | 225 | 345 | 1942 | 1948 | 129 | 234 | 1183 | 207 |
| **G11** | | 2059 | 17 | 142 | | 17 | 145 | 303 | | | 14 | 2 |
| **EXCLUDE** | 625 | 444 | | 138 | 147 | 193 | 34 | 69 | 60 | 214 | 1074 | 111 |

*SOM with single largest layout and complete dataset*

| | G1 | G2 | G3 | G4 | G5 | G6 | G7 | G8 | G9 | G10 | G11 | EXCLUDE |
|---|---|---|---|---|---|---|---|---|---|---|---|---|
| **G1** | 2127 | 1390 | 3456 | 517 | 92 | 1614 | | 5279 | | 276 | 9941 | 1012 |
| **G2** | | | | | 140 | 47 | 159 | | | | | |
| **G3** | 3636 | 6 | 47 | 10 | 10 | 2583 | 209 | 2 | | 106 | 235 | 51 |
| **G4** | 1126 | | | | | 62 | 48 | | | | | |
| **G5** | 82 | 240 | | | 98 | 339 | 1219 | | 5 | 99 | | |
| **G6** | | | | | | | 501 | | | | | |
| **G7** | 479 | 3 | 1 | | 3 | 13 | | 34 | | | 336 | |
| **G8** | | | | | 263 | 862 | | | 265 | 16 | | |
| **G9** | | | | | | 52 | 1 | | | | | |
| **G10** | 1876 | 513 | 382 | 144 | 1631 | 2097 | 162 | 848 | 1739 | 1701 | 1142 | 197 |
| **G11** | 1966 | 1 | 2 | 9 | 119 | 297 | 3 | 5 | 16 | 265 | 14 | 2 |
| **EXCLUDE** | 439 | 283 | 375 | 68 | 26 | 136 | | 564 | | 67 | 1051 | 100 |

*SOM with single largest layout and subset*